\documentclass[preprint,12pt,authoryear]{elsarticle}

\usepackage{natbib}
\setcitestyle{numbers,sort&compress}
\setcitestyle{square}

\usepackage[margin=1in]{geometry}
\usepackage{amsmath}
\usepackage{amsfonts}
\usepackage{amssymb}
\usepackage{bbm}
\usepackage{stackengine}
\usepackage{xcolor}
\usepackage{algorithm}
\usepackage{algpseudocode}
\usepackage{listings}
\usepackage{booktabs}
\usepackage{epsfig}
\usepackage{color}
\usepackage{graphicx}
\usepackage{float}
\usepackage{hyperref}
\usepackage{footnote}
\usepackage{mathrsfs}
\usepackage{booktabs}
\usepackage{listings}
\usepackage{multirow}
\usepackage{caption}
\usepackage{subcaption}
\usepackage{threeparttable}
\usepackage{diagbox}
\usepackage[section]{placeins}
\usepackage{soul}
\usepackage[misc,geometry]{ifsym}

\newtheorem{proposition}{Proposition}
\newtheorem{lemma}{Lemma}
\newproof{proof}{Proof}

\begin{document}

\begin{frontmatter}

\title{Deep Convolutional Neural Networks 
  Meet Variational Shape Compactness Priors for Image Segmentation}

\author[inst1]{Kehui Zhang}
\author[inst2]{Lingfeng Li}
\author[inst1]{Hao Liu}
\author[inst3]{Jing Yuan}
\author[inst4]{Xue-Cheng Tai \Letter}

\affiliation[inst1]{organization={Department of Mathematics, Hong Kong Baptist University},%Department and Organization
            addressline={Kowloon Tong}, 
            city={Hong Kong}, 
            country={China}}

\affiliation[inst2]{organization={Hong Kong Center for Cerebro-Cardiovascular Health Engineering},%Department and Organization
            addressline={Sha Tin}, 
            city={Hong Kong}, 
            country={China}}

\affiliation[inst3]{organization={College of Mathematical Medicine, Zhejiang Normal University},%Department and Organization
            addressline={Jinhua}, 
            city={Zhejiang}, 
            country={China}}

\affiliation[inst4]{organization={Norwegian Research Center},%Department and Organization
            city={Bergen}, 
            country={Norway}
            }

\journal{Neural Networks}

\begin{abstract}
Shape compactness is a key geometrical property to describe interesting regions in many image segmentation tasks. In this paper, we propose two novel algorithms to solve the introduced image segmentation problem that incorporates a shape-compactness prior. Existing algorithms for such a problem often suffer from computational inefficiency, difficulty in reaching a local minimum, and the need to fine-tune the hyperparameters. To address these issues, we propose a novel optimization model along with its equivalent primal-dual model and introduce a new optimization algorithm based on primal-dual threshold dynamics (PD-TD). Additionally, we relax the solution constraint and propose another novel primal-dual soft threshold-dynamics algorithm (PD-STD) to achieve superior performance. Based on the variational explanation of the sigmoid layer, the proposed PD-STD algorithm can be integrated into Deep Neural Networks (DNNs) to enforce compact regions as image segmentation results.
% Shape compactness is a key geometrical property to describe interesting regions in many image segmentation tasks. In this paper, we propose two novel algorithms to solve a  proposed  image segmentation model that incorporates a shape-compactness prior. Existing algorithms for such a problem often suffer from computational inefficiency, difficulty in reaching a local minimum, and the need to fine-tune the hyperparameters. To address these issues, we propose a novel optimization model along with its equivalent primal-dual model and introduce a as a min-max problem by dual expressions, and introduce a simple but new optimization algorithm based on primal-dual threshold dynamics (PD-TD).
% Additionally, we relax the solution constraint and propose another novel primal-dual soft threshold-dynamics algorithm (PD-STD) to achieve superior performance. Based on the variational explanation of the sigmoid layer, the proposed PD-STD algorithm for shape compactness prior can be integrated into Deep Neural Networks (DNNs) which guarantees that the output objects from the new designed neural networks are shape compact.
% and ensure segmentation results of the neural networks with compact shapes. 
Compared to existing deep learning methods, extensive experiments demonstrated that the proposed algorithms outperformed state-of-the-art algorithms in numerical efficiency and effectiveness, especially while applying to the popular networks of DeepLabV3 and IrisParseNet with higher IoU, dice, and compactness metrics on noisy Iris datasets. In particular, the proposed algorithms significantly improve IoU by $20\%$ training on a highly noisy image dataset.
\end{abstract}

\begin{keyword}
%% keywords here, in the form: keyword \sep keyword
Shape compactness \sep Image segmentation \sep Deep neural networks \sep threshold dynamics 
%% PACS codes here, in the form: \PACS code \sep code
% \PACS 0000 \sep 1111
%% MSC codes here, in the form: \MSC code \sep code
%% or \MSC[2008] code \sep code (2000 is the default)
% \MSC 0000 \sep 1111
\end{keyword}

\end{frontmatter}

\section{Introduction}

Image segmentation is a fundamental process in computer vision that involves partitioning a digital image into distinct non-overlapping subregions, including a wide spectrum of real-world applications such as autonomous driving \cite{mattyus2017deeproadmapper}, medical image segmentation \cite{kamnitsas2016deepmedic,UNet,milletari2016v}, and satellite image processing \cite{kampffmeyer2016semantic,pan2020deep}. In this respect, the variational method often provides a popular and mathematically explanable approach by formulating image segmentation as a minimization model whose energy function usually contains a data fidelity term and a boundary regularization term. The often-used variational models include the Mumford-Shah model \cite{mumford1989optimal}, the Chan-Vese model \cite{chan2001active}, and the Potts model \cite{potts1952some}. However, when the image is corrupted by noise or the object of interest is partially blocked in the image, such variational models may not produce desirable segmentation results; hence, some task-dependent prior knowledge is incorporated into the variational model by adding extra penalization functions or hard constraints, so as to improve the final results. Many shape priors have been studied in the literature, such as the convexity prior \cite{LiuConvex}, the shape volume prior \cite{liu2022deep}, the star-shape prior \cite{veksler2008star}, and the compactness prior \cite{TRIC,dolz2017unbiased}, etc. 

In this work, we study the variational image segmentation model with a compactness shape prior. 
The shape compactness prior helps to obtain a more compact segmentation mask with smooth boundaries in the results, by introducing an additional penalization term to describe a compact region geometrically~\cite{TRIC,dolz2017unbiased}. Previous studies have explored different definitions of shape compactness. 
The perimeter-to-area ratio was proposed in \cite{Stereo} to depict the compactness of the segmented regions.  
The authors in \cite{TRIC} introduced a new compactness term that is invariant from scale, rotation, and translation. 
The authors of \cite{dolz2017unbiased} first integrate compactness, in terms of the ratio of the squared perimeter to the area, into image segmentation to ensure size invariance. In addition, it is well-known  that the ratio of squared perimeter to area achieves the minimum when the shape is circular, which implies its preference to circular shape segments in the result.
However, adding this ratio leads to a challenging non-convex optimization problem. The authors of \cite{dolz2017unbiased} proposed an algorithm based on the alternating-direction method of multipliers (ADMM), which is, however, computationally expensive and requires fine-tuning of many hyper-parameters.

% Training deep networks is often time-consuming, but its evaluation is very fast. 
Given the great success of deep learning-based (DL-based) methods in many different applications during the past decades, many neural network architectures have been proposed for image processing, such as LeNet \cite{lecun1989handwritten}, AlexNet \cite{krizhevsky2012imagenet}, and U-Net \cite{UNet}, etc. 
In a series of recent works \cite{DeepLabV1,DenseASPP,ENet}, dilated convolutions, which increase the receptive field without a significant increase in computational cost, are used to achieve state-of-the-art segmentation performance. 
Specifically, U-Net, as proposed in \cite{UNet}, employs an encoder-decoder structure that concatenates multiscale features, allowing effective image segmentation. 
Such an architecture has inspired several works, such as SegNet \cite{SegNet}, U-Net++ \cite{zhou2019unet++}. 
In \cite{tai2023pottsmgnet}, a novel neural network framework, the so-called PottsMGNet, is proposed to solve the variational Potts model of image segmentation by leveraging operator-splitting methods, multigrid methods, and control variables. 
Moreover, PottsMGNet shows that most encoder-decoder-based networks are equivalent to some optimal control problems  for certain initial value problems using operator-splitting methods as solvers.
Although deep learning (DL)  based methods have achieved remarkable performance for image segmentation, integrating spatial information priors into such DL-based methods still remains a challenge. 
Recently, the combination of shape priors with deep learning models was explored such that semantic features were automatically extracted from large datasets using neural networks. 
Together with the Potts model, operator splitting methods, and double-well potential, the U-Net architecture is used in \cite{liu2023double} to segment images with length penalties. 
In \cite{liu2022deep}, a method is proposed to incorporate spatial priors by introducing the variational explanation of the sigmoid function. This method allows for the integration of shape priors, such as convexity and star shape, into deep learning-based approaches. Motivated by this, this paper aims to incorporate compactness priors with DL-based methods.

In this paper, we consider a non-convex optimization model for image segmentation with the shape compactness prior, for which we introduce two primal-dual-based (PD-based) optimization algorithms using threshold dynamics (TD) and soft threshold dynamics (STD), respectively. The proposed algorithms enjoy advantages in both theory and numerical solvers. 
% Compared to the ADMM-based algorithm \cite{dolz2017unbiased}, the proposed algorithms are easier to implement and perform more efficiently and accurately. 
Moreover, the introduced PD-STD algorithm can be easily integrated into many popular neural networks such as  DeepLabV3 and IrisParseNet, and significantly improves performance while segmenting noisy images.
% To demonstrate its effectiveness, we will integrate our method with some network structures such as DeepLabV3 and IrisParseNet. As shown in Figure \ref{fig:real} - \ref{fig:fundus}, our algorithms significantly improve performance and robustness when dealing with noisy testing datasets.  
We list our main contributions of this work as follows:
\begin{itemize}
    \item []- We study Potts model joint with a shape compactness prior for image segmentation, and introduce its novel equivalent primal-dual and dual models. % which lead to new algorithms that are efficient and  can easily implemented. 
    \item []- Two novel optimization algorithms, namely the PD-TD (Primal-Dual Threshold Dynamics) and the PD-STD (Primal-Dual Soft Threshold Dynamic) algorithms, are proposed based on the introduced primal-dual and dual models, which outperform previous methods in efficiency and accuracy.
    % to solve the model. Compared to the previous method, the proposed methods are more efficient and more accurate.
    \item []- Motivated by \cite{liu2022deep}, the PD-STD algorithm can be integrated into the structures of deep neural networks. The new networks demonstrate better robustness for segmenting objects of circular shapes, especially on noisy images.  
\end{itemize}
% which is related to the softmax layer in CNN to solve the model.  It will give us better results and this layer can be added after any backbone to replace the activation function to get more circular results.

This work is organized as follows. In Sec. \ref{sec:Preliminaries}, we discuss related works of the classical variational segmentation models, especially the models with shape-compactness prior; also, recent deep learning-based image segmentation methods. 
We introduce a new variational model in Sec.~\ref{sec: Proposed Method}, and propose its novel equivalent models and corresponding algorithms in Sec. \ref{sec: Proposed Algorithms}. In Sec. \ref{sec: Numerical experiments}, experimental results are presented to verify the efficiency, accuracy, and robustness of our proposed algorithms. 
The conclusions will be given in Sec. \ref{sec: Conclusion}.

\section{Preliminaries} \label{sec:Preliminaries}
\subsection{Variational segmentation models} \label{subsection: variModels}
% Suppose $\Omega$ is a compact domain in $\mathbb{R}^2$.
% The objective of segmentation is to divide the input image $v(x): \Omega \to \mathbf{R}$ into multiple distinct phases denoted by $\Omega_i$, where $i$ ranges from $1$ to $N$. The aim is to assign each pixel in the image to a specific phase.  One of the extensively researched image segmentation models is the Potts model \cite{potts1952some}, which originated from statistical mechanics to represent interactions between spins on a crystalline lattice. Over time, its utility has been recognized in computer vision and signal processing, leading to its application in discrete optimization by researchers such as Geman \cite{geman1984stochastic} and Boykov et al. \cite{boykov1998markov,Boykov01fastapproximate}. In recent times, the continuous variational extension of the Potts model has gained significant popularity:
Suppose $\Omega$ is an input compact image domain in $\mathbb{R}^2$.
The objective of binary image segmentation is to divide the input image $\mathcal{P}(x): \Omega \to \mathbb{R}$ into two distinct regions, i.e. the foreground region $\Omega_0$ and the background region $\Omega \setminus \Omega_{0}$. 
% $\Omega_0$ is the specific region within $\Omega$ we aim to segment. $\Omega \setminus \Omega_{0}$ represents the background. 
The aim is to assign each pixel $x\in\Omega$ to a specific region of the foreground and background, respectively. 
One of the most popular image segmentation models is the Potts model \cite{potts1952some}, which originated from statistical mechanics to represent interactions between spins on a crystalline lattice and well studied in a discrete optimization setting by Geman \cite{geman1984stochastic} and Boykov et al. \cite{boykov1998markov,Boykov01fastapproximate}. 
% Over time, its utility has been recognized in computer vision and signal processing, leading to its application in discrete optimization by researchers such as Geman \cite{geman1984stochastic} and Boykov et al. \cite{boykov1998markov,Boykov01fastapproximate}. 
During recent decades, the extension of Potts model in the continuous optimization setting has gained significant attentions in image segmentation, which aims to minimize the following energy function:
% \begin{align}
% \label{eq: potts_ori}
%     \min_{\{\Omega_i\}_{i=1}^{N}} E_{Potts} (\{\Omega_i\}) &= \sum_{i=1}^{N} \int_{\Omega_i} f_i(x) \, dx + \alpha \sum_{i=1}^{N} |\partial \Omega_i|,\\
%     s.t. \quad \bigcup_{i=1}^{N} \bar{\Omega_i} &= \Omega,\\
%     \Omega_i \cap \Omega_j &= \emptyset, \quad \text{ $\forall$ i$\neq$ j, \quad i,j $\in$ [1, N]}.
% \end{align}
\begin{align}
\label{eq: potts_ori}
    \min_{\Omega_0} E_{\mathrm{Potts}} (\{\Omega_0\}) &:= \int_{\Omega_0} f(x) \, dx + \alpha  |\partial \Omega_0|,
\end{align}
where $f(x)$ is the data ﬁdelity term at each pixel $x\in \Omega$, depending on the given image information $\mathcal{P}(x)$, and the second term $|\partial \Omega_0|$, named edge force term, measures the boundary length of the foreground region $\Omega_0$. Clearly, the second term serves as the regularization term promoting smooth and well-defined boundaries.

% One classic implementation of the Potts model is proposed in \cite{lellmann2009convex,zach2008fast} and it uses the simplex-constrained vector functions. 
Let $u(x)$ be the discrete-valued labeling function $u: \Omega \to S$ such that
% We can define a vector-valued labeling function $u: \Omega \to S$ where the constraint set is defined as 
% \begin{align}
%     S = \left\{\mathbf{u}(x) = (u_1(x), u_2(x), \dots, u_N(x)) \in \{0, 1\}^N \bigg|\sum_{i=1}^N u_i(x) = 1, \forall x \in \Omega \right\},
% \end{align}
\begin{align}
    S = \left\{{u}(x) \in \{0, 1\}, \forall x \in \Omega\bigg|u\in L^1(\Omega) \right\}, \quad \text{where}\;\; u(x) = 
\begin{cases}
    1, x\in \Omega_0,\\
    0, x \notin \Omega_0.
\end{cases}
\end{align}
% and
% $$
% u(x) = 
% \begin{cases}
%     1, x\in \Omega_0,\\
%     0, x \notin \Omega_0.
% \end{cases}
% $$
The optimization model \eqref{eq: potts_ori} can, therefore, be rewritten as
\begin{align}
    \label{eq: strictPotts}
    \min_{{u}(x) \in S} \int_{\Omega} f(x)u(x)\, dx + \alpha \int_{\Omega} |\nabla u(x)|\, dx  , 
\end{align}
where the total-variation function gives the estimation of the perimeter of the foreground region, i.e. $|\partial \Omega_0|$, in a continuous space setting.

Given the discrete-valued labeling function $u(x)$, the optimization problem (\ref{eq: strictPotts}) is non-convex, thus difficult to solve directly. To address this, its convex relaxation is thus studied:
\begin{align}
    \label{eq: relaxPotts}
    \min_{{u} \in \mathcal{K} }  \int_{\Omega} f(x)u(x)\, dx + \alpha \int_{\Omega} |\nabla u(x)|\, dx,   
\end{align}
where 
% \begin{align}
% \label{set:tildeS}
%     \Tilde{S} = \left\{u(x)\in [0, 1], \forall x \in \Omega \right\}.
% \end{align}
\begin{align}
\label{def: setK}
    \mathcal{K}=\left\{u(x)\in[0,1],\forall x\in\Omega\bigg|u\in L^1(\Omega)\right\}.
\end{align}
It can be proved that the global optimum of the non-convex discrete optimization problem can be easily obtained by simply thresh-holding the optimum of its convex relaxation problem \cite{chan2006algorithms,yuan2010continuous}. 
Multiple efficient algorithms have been developed to address the convex optimization problem \eqref{eq: relaxPotts}, based on primal-dual optimization \cite{ChPo110a,EsZhFc100a,WT10} and dual optimization \cite{yuan2010continuous} respectively. More details about literature proposed  algorithms for solving the Potts model and relevant references can be found in a recent survey paper \cite{tai2021potts}.
% One method involves using total variation to transform the Potts model into a min-max problem thereby formulating it as a primal-dual problem. The resulting primal-dual model can be effectively solved using the primal-dual algorithm \cite{ChPo110a,EsZhFc100a,WT10}. Another method to address the Potts model entails proposing a continuous max-flow model \cite{yuan2010continuous} based on the simplex constrained Potts model (\ref{eq: strictPotts}). To address this formulation, an equivalent primal-dual formulation is introduced by incorporating Lagrangian multipliers. The resulting problem can be efficiently solved using the ADMM algorithm. 
% More details of the implementations of Potts models can be found in \cite{tai2021potts}.

\subsection{Shape compactness prior}
% Suppose we are dealing with the binary segmentation tasks and $\Omega_0$ is the specific region within $\Omega$ we aim to segment. 
% Variational segmentation models are designed to partition an image into coherent regions through the optimization of an energy functional. 
% To guide the segmentation process and maintain spatial regularity, we often incorporate shape prior information into segmentation models. By confining the segmentation to specific spaces, the accuracy and robustness of the technique are enhanced. 
In practice, shape prior information is often incorporated into the applied variational image segmentation models so as to further constrain the solution space of segmentation and improve accuracy and robustness of computation results.
Various types of shape prior, such as convexity and star-shape, are proposed and explored in variational image segmentation models \cite{liu2022deep,LiuConvex}. 
% {\color{red} [references]!}.
It is of great interest to segment the compact target regions from the given images in many real-world image segmentation tasks. 
The key challenges lie in how to represent such region compactness in the most effective and descent way in mathematics and solve it efficiently in numerics. 
This is one of the main research topics in this study.
% In this study, we focus on integrating such region compactness information into the variational image segmentation model studied. 
% to generate more compact results. The key challenge lies in determining the most effective representation of compactness within the model.

% With the domain $\Omega$ and the region of interest $\Omega_0$ defined in Section \ref{subsection: variModels}, we can define the shape compactness of $\Omega_0$ as follows:
% $$\frac{\text{Per}^2(\Omega_0)}{\text{Area}(\Omega_0)}.$$
Given the isoperimetric inequality \cite{osserman1978isoperimetric} for the region $\Omega_0$, we have
$$ \text{Per}(\Omega_0)\geq 2\sqrt{\pi\text{Area}(\Omega_0)} $$ 
and the equality holds if and only if $\Omega_0$ is a ball.
Clearly, the ratio of $\text{Per}^2(\Omega_0)$ and $\text{Area}(\Omega_0)$ is always larger than or equal to $4\pi$, and the minimum is obtained if and only if $\Omega_0$ is a ball. Its minimization simply enforces the result of segmentation to be a single compact region. 
Therefore, we define the following formulation as the measure of shape compactness for the region $\Omega_0$:
$$\frac{\text{Per}^2(\Omega_0)}{\text{Area}(\Omega_0)}.$$

% Therefore, the shape compactness measure is always larger than or equal to $4\pi$, and the minimum is obtained if and only if $\Omega_0$ is a ball.

Let $u(x): \Omega \to \mathbb{R}$ be the indicator function of the region $\Omega_0\subset\mathbb{R}^2$:
$$ u(x)=\begin{cases}1, & x \in \Omega_0,  \\
0, & x \notin \Omega_0 .\end{cases} $$
It is well-known that, for the perimeter of $\Omega_0$, we have $\text{Per}(\Omega_0)=|u|_{TV}$ \cite{chambolle2010introduction}, where $|u|_{TV}$ is the total-variation of $u(x)$ such that
$$ |u|_{TV}=\sup_{v\in\mathcal{V}}-\int_{\Omega}u\text{div}(v)dx, $$
where
$$\mathcal{V}=\{v\in \left(C_0^1(\Omega)\right)^2, |v(x)|\leq 1\ \forall x\in\Omega \}. $$
% The area of $\Omega_0$ can be measured by integrating $u$ over $\Omega$: $\int_{\Omega}u(x)dx$ and the perimeter of $\Omega_0$ can be measured by the total variation (TV) of $u$:
% $$ |u|_{TV}=\sup_{v\in\mathcal{V}}-\int_{\Omega}u\text{div}(v)dx $$
% where
% $$\mathcal{V}=\{v\in \left(C_0^1(\Omega)\right)^2, |v(x)|\leq 1\ \forall x\in\Omega \}. $$
% A well-known property is that $|u|_{TV}=\text{Per}(\Omega_0)$ \cite{chambolle2010introduction}.

Thus, the shape compactness of $\Omega_0$ can be properly reformulated as 
\begin{equation}\label{shapecptmetric}
\frac{(|u|_{TV})^2}{\int_{\Omega}u(x)dx}.
\end{equation}
The area of $\Omega_0$ is assumed to be non-zero in this paper.

\subsection{Deep learning-based segmentation models}
% We define the input image as $\boldsymbol{v}^0 = \boldsymbol{v}$, which serves as the input for a Deep Neural Network (DCNN) designed for pixel-wise segmentation. 
% \textbf{say some words before defining DNN}

This work also introduces the incorporation of the shape compactness prior, as defined in \eqref{shapecptmetric}, into modern neural networks, such as DNNs. DNNs emerged as a widely used tool in the field of artificial intelligence and have revolutionized numerous domains ranging from computer vision to natural language processing. 
Especially, with their great ability to learn hierarchical representations of data through multiple layers of interconnected neurons, DNNs have achieved remarkable success in capturing complex patterns and obtaining state-of-the-art performance in various challenging tasks, and emerged as a widely used tool in the field of artificial intelligence. 
% With their great ability to learn hierarchical representations of data through multiple layers of interconnected neurons, DNNs have achieved remarkable success in capturing complex patterns and obtaining state-of-the-art performance in various challenging tasks. 
We review the often-used structure of DNNs and discuss the motivation of integrating shape prior information into DNNs.

DNNs for image segmentation can be expressed as a parametrized nonlinear operator $\mathcal{N}_{\boldsymbol{\Theta}}(\boldsymbol{v}^{0})$ for a multilayer neural network, where $\boldsymbol{v}^{0}$ denotes the input image to the network and its output of each network layer is denoted by $\boldsymbol{v}^t$, $t=1, ..., T$. The structures of neural networks are typically compositions of such layers, e.g. DeepLabV3 \cite{DeeplabV3}:
\begin{equation}
\begin{cases}
\label{eq: DCNNs}
    \boldsymbol{o}^t = \mathcal{T}_{\boldsymbol{\Theta}^{t-1}}(\boldsymbol{v}^{t-1}, \boldsymbol{v}^{t-2}, \dots, \boldsymbol{v}^{0}),\  t=1, 2, \dots, T,\\
    \boldsymbol{v}^{t} = \mathcal{A}^t(\boldsymbol{o}^{t}), t=1, 2, \dots, T.
\end{cases}
\end{equation}
In \eqref{eq: DCNNs}, the activation function $\mathcal{A}^t$ can be chosen as many different operators, such as ReLU, sigmoid, and tanh. It can also include downsampling, upsampling operators, and their compositions. In the final layer, $\mathcal{A}^t$ is typically a soft classification activation function of sigmoid or softmax.
The operator $\mathcal{T}_{\boldsymbol{\Theta}^{t-1}}(\boldsymbol{v}^{t-1}, \boldsymbol{v}^{t-2}, \dots, \boldsymbol{v}^{0})$ describes the connections between the t-th layer $\boldsymbol{v}^{t}$ and its previous layers $\boldsymbol{v}^{t-1}, \boldsymbol{v}^{t-2}, \dots, \boldsymbol{v}^{0}$.   
For example, in a simple convolutional network, $\boldsymbol{v}^{t}$ is only associated to its previous layer $\boldsymbol{v}^{t-1}$ and $\mathcal{T}_{\boldsymbol{\Theta}^{t-1}}(\boldsymbol{v}^{t-1}) = \boldsymbol{\omega}^{t-1} * \boldsymbol{v}^{t-1} + \boldsymbol{b}^{t-1}$ where $\boldsymbol{\omega}^{t-1}$ and $\boldsymbol{b}^{t-1}$ are the convolution kernel and the bias of an affine transformation, respectively.  
Let $\boldsymbol{\Theta}$ be the collection of learnable parameters:
$$\boldsymbol{\Theta} = \left\{ \boldsymbol{\Theta}^{t} = (\boldsymbol{\omega}^{t-1}, \boldsymbol{b}^{t-1}) \bigg| t=0, 1, \dots, T-1  \right\}.$$
% By making careful choices for the operator $\mathcal{A}^t$, such as using ReLU, downsampling, or upsampling operators, and allowing $\mathcal{T}_{\boldsymbol{\Theta}^{t-1}}$ to establish connections between different layers, the formulation (\ref{eq: DCNNs}) can effectively represent the well-known backbones such as DeepLabV3 \cite{DeeplabV3}, SegNet\cite{SegNet}, U-Net\cite{UNet}.  
%??? add all the other popular networks. ??? 
% It becomes apparent that the operators $\mathcal{A}^t$ and $\mathcal{T}_{\boldsymbol{\Theta}^{t-1}}$ exhibit continuity properties, including the possibility of being Lipschitz continuous. Consequently, the overall operator $\mathcal{N}_{\boldsymbol{\Theta}}$ is continuous.

For the problem of binary image segmentation, the sigmoid function is often chosen as the activation function of the final layer $\mathcal{A}^T$ in DNNs:
\begin{align}
    S(o^{T})(x) = \frac{1}{1+e^{-o^{T}(x)}},
\end{align}
where $o^{T}: \Omega \to \mathbb{R}$ is the logits output of the neural network, $S(o^{T})$ maps the logits from $(-\infty, +\infty)$ to $[0, 1]$ and enforces output probabilities.
%For multi-class segmentation, the softmax function can be used.  

However, integrating high-level spatial priors into DNNs is still a challenge for the often-used DNNs. 
In this respect, Liu et al \cite{LiuConvex} proposed an interesting work to naturally enforce the classical spatial regularization term into DNNs:
\begin{equation}
\begin{cases} \label{eq: DNNwithRegularization}
    {o}^t = \mathcal{T}_{\boldsymbol{\Theta}^{t-1}}({v}^{t-1}, {v}^{t-2}, \dots, {v}^{0}),\  t=1, 2, \dots, T,\\
    {v}^{t} = \mathcal{A}^t({o}^{t}), t=1, 2, \dots, T-1,\\
    {v}^T = \arg\min_{u\in \mathcal{K}} \left\{\lambda\langle -{o}^T,{u}\rangle+\epsilon\langle {u},\log({u})\rangle + \epsilon\langle 1-{u},\log(1-{u})\rangle + \mathcal{R}({u}) \right\}.
\end{cases}
\end{equation}
It is motivated by the fact that the sigmoid function can be regarded as the minimizer of a dual function of the entropy-regularized Rectified Linear Unit (ReLU):
\begin{align}
    \label{eq: variational explanation}
    \arg\min_{u \in \mathcal{K}} \left\{ <-o, u> + \epsilon <u, \log (u)> + \epsilon <1-u, \log (1-u)> \right\},
\end{align}
when the entropic regularization parameter $\epsilon = 1$. 
Clearly, the last layer of \eqref{eq: DNNwithRegularization} just represents a total-variation regularized sigmoid function.

Inspired by \cite{LiuConvex}, we propose to incorporate the {studied shape-compactness prior into DNNs, to push the spatial compactness of the segmentation result from DNNs. 

\section{Image segmentation with compactness prior} \label{sec: Proposed Method}
In this work, let $C(u)$ be the spatial compactness regularization of a function $u(x)\in\mathcal{S}$, where $\mathcal{S}$ is defined in Sec. \ref{subsection: variModels}, as below: 
$$ C(u)=\begin{cases}
4\pi, & u(x)\equiv 0,\\
(|u|_{TV})^2/\int_{\Omega}u(x)dx,& \text{otherwise}.
\end{cases}
$$
Given the classical image segmentation model \eqref{eq: strictPotts}, we consider replacing its length-minimization term, i.e. the total-variation function, by the above shape-compactness prior, and propose the new image segmentation model as follows:
%Since we want to help the model generate more compact segmentation results, we may consider replacing the regularization term in equation (\ref{eq:potts}) with the shape compactness regularization term. By incorporating this term, the modified model can be expressed as follows:
\begin{align}
    \min_{u\in \mathcal{S}} E_\lambda(u):=\lambda\langle f,u\rangle+C(u), \label{eq:primal1}
\end{align}
where the region force $f(x) \in L^\infty(\Omega)$ and $\langle\cdot,\cdot\rangle$ denotes the $L^2$ inner product.

\subsection{Analysis of the optimization model \eqref{eq:primal1}} \label{sec:model}

We first show that the minimizer of the novel shape-compactness regularized image segmentation model \eqref{eq:primal1} does exist and its minimizer tends to be a circular region when the regularization parameter in \eqref{eq:primal1} becomes small enough.
\begin{proposition}\label{lemma:lsc}
Suppose $\{u_n\}_{n=1}^\infty$ is a sequence of functions in $\mathcal{S}$ and $u_n\rightarrow u^*$ in $L^1(\Omega)$. Then $C(u^*)\leq \liminf_{n\rightarrow\infty}C(u_n).$ 
\end{proposition}
\begin{proof}
Since $\mathcal{S}$ is closed, $u^*$ is also in $\mathcal{S}$.
If $u^*=0$, then $C(u^*)=4\pi\leq\liminf_{n\rightarrow\infty}C(u_n)$ follows directly from the isoperimetric inequality. If $u^*\neq 0$, then, by the lower semi-continuous of total variation \cite{acar1994analysis}, we have $\lim_{n\rightarrow\infty}\int_{\Omega}u_ndx=\int_{\Omega}u^*dx$ and  $|u^*|_{TV}\leq\liminf_{n\rightarrow\infty}|u_n|_{TV}$. Thus,
$C(u^*)\leq \liminf_{n\rightarrow\infty}C(u_n).$
$\hfill\square$
\end{proof}
\begin{lemma}[Existence of minimizer]\label{lemma:existence}
For any $\lambda>0$, there exists a minimizer of $E_\lambda(u)$ in $\mathcal{S}$.
\end{lemma}
\begin{proof}
Suppose $\{u_n\}_{n=1}^\infty$ is a minimizing sequence of $E_\lambda$ in $\mathcal{S}$ such that $\lim_{n\rightarrow\infty}E_\lambda(u_n)=\inf_{v\in\mathcal{S}}E_\lambda(v)$. Then, there exists a constant $M$ such that
\begin{align*}
&M\geq E_\lambda(u_n)\geq -\lambda\Vert f\Vert_{\infty}|\Omega|+\frac{(|u_n|_{TV})^2}{|\Omega|}\\ 
\Rightarrow &|u_n|_{TV}\leq \sqrt{M|\Omega|+\lambda\Vert f\Vert_{\infty}|\Omega|^2}.
\end{align*}
In addition, we have $\Vert u_n\Vert_{L^1}\leq |\Omega|$. Thus, the BV norm of $u_n$ is uniformly bounded for $n=1,2,\dots$:
$$ \Vert u_n\Vert_{BV}=\Vert u_n\Vert_{L^1}+|u_n|_{TV}. $$
 By Theorem 2.5 of \cite{acar1994analysis}, there exists a sub-sequence $\{u_{n_k}\}_{k=1}^\infty$ converging to a function $u^*$ in $L^1(\Omega)$. Since $\mathcal{S}$ is closed, $u^*$ also belongs to $\mathcal{S}$.
By Proposition \ref{lemma:lsc}, we have $C(u^*)\leq \liminf_{k\rightarrow\infty}C(u_{n_k})$ for any $u^*\in\mathcal{S}$.
Then, 
$$\inf_{v\in\mathcal{S}}E_\lambda(v)\leq E_\lambda(u^*)\leq\liminf_{k\rightarrow\infty}E_\lambda(u_{n_k})=\inf_{v\in\mathcal{S}}E_\lambda(v).$$
Thus, $\inf_{v\in\mathcal{S}}E_\lambda(v)=E_\lambda(u^*)$ and $u^*\in\mathcal{S}$ is a minimizer.
$\hfill\square$
\end{proof}

\begin{lemma}[Convergence of minimizers]\label{lemma:convergence}
Denote $u_n:=\arg\min_{v\in\mathcal{S}}E_{1/n}(v)$, then every cluster point of the sequence $\{u_n\}$ is the indicator function of a ball in $\Omega$.
\end{lemma}
\begin{proof}
Let $\tilde{u}$ be the indicator function of a ball in $\Omega$, then 
$$-\frac{1}{n}\Vert f\Vert_\infty|\Omega|+4\pi\leq E_{1/n}(u_n)\leq E_{1/n}(\tilde{u})\leq \frac{1}{n}\Vert f\Vert_\infty|\Omega|+4\pi$$ for $n=1,2,\dots$. Thus, $\lim_{n\rightarrow\infty}E_{1/n}(u_n)=\lim_{n\rightarrow\infty}C(u_n)=4\pi$. Using a similar argument with Lemma \ref{lemma:existence}, we can show $\{u_n\}$ has bounded BV norm and thus converges to some $u^*\in\mathcal{S}$ up to a sub-sequence $\{u_{n_k}\}$. Then, $4\pi\leq C(u^*)\leq\liminf_{n\rightarrow\infty}C(u_{n_k})=4\pi$. Therefore, $C(u^*)=4\pi$ and $u^*$ must be the indicator function of a ball.
$\hfill\square$
\end{proof}
As Lemma \ref{lemma:convergence}, it is expected that the optimum of the proposed image segmentation problem (\ref{eq:primal1}) tends to be circular while the penalty parameter $\lambda$ is small enough. 

\subsection{Dolz-Ayed-Desrosiers algorithm\label{sec:admm}}
Dolz et al \cite{dolz2017unbiased} proposed an algorithm to solve the model \eqref{eq:primal1} based on the alternating direction methods of multipliers (ADMM). By the two auxiliary variables $s=\int_{\Omega }u(x)dx$ and $z=u$, the studied optimization model \eqref{eq:primal1} can be written as the following linear-equality constrained optimization problem:
\begin{align*}
    \min_{u\in\mathcal{S},z,s} & \left\{\lambda\langle f,u\rangle+\frac{|u|_{TV}|z|_{TV}}{s}\right\}, \\
    \text{s.t } & s=\langle z,\mathbbm{1}_{\Omega}\rangle,\quad u=z,
\end{align*}
where $\mathbbm{1}_{\Omega}$ is a constant function in $\Omega$ with all values equal to $1$. 
Its corresponding augmented Lagrangian function is:
\begin{align*}
    L(u,z,s,\nu_1,\nu_2)=&\lambda\langle f,u\rangle+\frac{|u|_{TV}|z|_{TV}}{s}+\frac{\mu_1}{2}\Vert u-z+\nu_1\Vert_2^2+\frac{\mu_2}{2}(s-\langle z,\mathbbm{1}_{\Omega}\rangle+\nu_2)^2.
\end{align*}
At each iteration, each variable of $(u,z,s)$ is updated separately while fixing the two multipliers of $(\nu_1,\nu_2)$ and the multipliers of $(\nu_1,\nu_2)$ are updated as the classical augmented Lagrangian method consequently. 
Particulalrly, the anisotropic total-variation functions of $|u|_{TV}$ and $|z|_{TV}$ are applied, so the step of $u-$update is solved by graph-cut and $z-$update is computed by tackling a large-scale linear equation; the step of $s-$update is simply finished by finding the root of a cubic equation.

\section{Novel equivalent models and efficient algorithms} \label{sec: Proposed Algorithms}
The algorithm proposed in \cite{dolz2017unbiased} has a series of hyper-parameters, which require fine-tuning and impacts its numerical performance in practice.
%A series of hyper-parameters require fine-tuning for the proposed algorithm in \cite{dolz2017unbiased} which impacts its numerical performance in practice.
%and is not computationally efficient. 
In this work, we reformulate the image segmentation model \eqref{eq:primal1} with the shape-compactness prior and propose new efficient algorithms.

\subsection{Novel equivalent optimization models}
% To develop fast algorithms for the model (\ref{eq:primal1}), we will first reformulate it.
As shown in \cite{liu2022deep,tai2023pottsmgnet}, the total-variation function $|u|_{TV}$, i.e. the boundary length term, can well approximated by
\begin{equation}
     \langle u,G_\sigma*(1-u)\rangle,\label{eq:primal}
\end{equation}
where $G_\sigma(x)=\frac{1}{2\pi\sigma^2}\exp(-\frac{|x|^2}{2\sigma^2})$ is the Gaussian kernel. 
Such an approximation $\Gamma$-converges to the exact length $|\partial \Omega|$ as $\sigma \to 0^+$. 

Hence, the approximated model of \eqref{eq:primal1} can be essentially formulated as
\begin{align}
    \min_{u\in \mathcal{S}}\left\{ \lambda\langle f,u\rangle+\frac{\langle u,G_\sigma*(1-u)\rangle^2}{\langle u,\mathbbm{1}_{\Omega}\rangle} \right\}. \label{eq:primal2}
\end{align}

Let $q = \langle u,G_\sigma*(1-u)\rangle$. In view of the following dual representation of the convex function $\frac{q^2}{\langle u,\mathbbm{1}_{\Omega}\rangle}$ such that
\[
\frac{q^2}{\langle u,\mathbbm{1}_{\Omega}\rangle} \,:=\, \max_{p \in \mathbb{R}} \left\{ p \cdot q - \frac{p^2\langle u,\mathbbm{1}_{\Omega}\rangle}{4}  \right\}\, ,
\]
the optimization model \eqref{eq:primal2} can be equally rewritten as 
\begin{align*}
    \min_{u\in \mathcal{S}}\max_{p\in\mathbb{R}} \left\{ \lambda\langle f,u\rangle+p\langle u,G_\sigma*(1-u)\rangle-\frac{p^2\langle u,\mathbbm{1}_{\Omega}\rangle}{4}\right\}. 
\end{align*}

By simple calculation, the above optimization problem becomes
\begin{align}
    \min_{u\in \mathcal{S}}\max_{p\in\mathbb{R}}\; L(u,p)\left\{:= 
    \langle\lambda f-\frac{p^2}{4},u\rangle+p \langle u,G_\sigma*(1-u)\rangle \right\} \, .
    \label{eq:primal-dual}
\end{align}
In this work, the min-max problem \eqref{eq:primal-dual} is called the equivalent \emph{primal-dual model} of \eqref{eq:primal2}.

\subsection{Primal-dual algorithms using threshold dynamics (PD-TD)}
% It has been proved that the boundary length $|u|_{TV}$ is proportional to \cite{ThresholdDynamics}:

% Based on the primal-dual formulation (\ref{eq:primal-dual}), we propose a simple primal-dual algorithm based on the threshold dynamics to update $u$ and $p$ separately, to find optimizers for this problem. Let's denote values of $u$ and $p$ at the $k$th iteration as $u^k$ and $p^k$. When updating $u^{k+1}$, we fix $p=p^k$ and solve
Given the primal-dual formulation (\ref{eq:primal-dual}), we propose the first primal-dual algorithm based on threshold dynamics.
Given $u^k$ and $p^k$ at the $(k+1)$-th step, the new algorithm involves two steps to update $u^{k+1}$ and $p^{k+1}$:

First, we fix $p^k$ and update $u$ by
\begin{align}
    u^{k+1}=\arg\min_{u\in\mathcal{S}}L(u,p^k), \label{eq:primal_update_TD}
\end{align}
which can be solved approximately by linearizing $\langle u,G_\sigma*(1-u)\rangle$ at $u^k$:
\begin{align*}
    \langle u,G_\sigma*(1-u)\rangle\approx \langle u^k,G_\sigma*(1-u^k)\rangle+\langle u-u^k,G_\sigma*(1-2u^k)\rangle . 
\end{align*}
The optimum $u^{k+1}$ can, therefore, be efficiently computed by threshholding $\phi^k(x)$, i.e.
\begin{align} \label{hth}
    u^{k+1}&= \mathbbm{1}_{\phi^k<0}=\begin{cases}
    1 & \phi^k(x)<0, \\
    0 & \phi^k(x)\geq 0,
    \end{cases}
\end{align}
where
$$\phi^k(x)=\lambda f(x)-\frac{(p^k)^2}{4}+p^kG_\sigma(x)*(1-2u^k(x)).$$

% \begin{align}
%     u^{k+1}=\arg\min_{u\in\mathcal{S}}L(u,p^k). \label{eq:primal_update_TD}
% \end{align}
% This sub-problem can not be solved explicitly, but we can approximate it by linearizing $\langle u,G_\sigma*(1-u)\rangle$ at $u^k$:
% \begin{align}
%     \langle u,G_\sigma*(1-u)\rangle\approx \langle u^k,G_\sigma*(1-u^k)\rangle+\langle u-u^k,G_\sigma*(1-2u^k)\rangle.
%     \label{eq:linearize}
% \end{align}
% Then, the $u$ update is given by
% \begin{align*}
%     u^{k+1}&=\arg\min_{u\in\mathcal{S}}L(u,p^k)\\
%     &\approx \arg\min_{u\in\mathcal{S}} \langle\phi^k,u\rangle\\
%     &=\mathbbm{1}_{\phi^k<0}=\begin{cases}
%     1 & \phi^k(x)<0 \\
%     0 & \phi^k(x)\geq 0
%     \end{cases}
% \end{align*}
% where
% $$\phi^k(x)=\lambda f(x)-\frac{(p^k)^2}{4}+p^kG_\sigma(x)*(1-2u^k(x)).$$

Second, we fix $u^{k+1}$ and update $p$ by solving a quadratic equation explicitly:
% When updating $p$, we just need to solve a quadratic equation, which has an explicit solution:
\begin{align} p^{k+1}=\arg\max_{p\in\mathbb{R}} L(u^{k+1},p)=\frac{2 \langle u^{k+1},G_\sigma*(1-u^{k+1})\rangle}{\langle u^{k+1},\mathbbm{1}_{\Omega}\rangle}. \label{eq:dual_update_TD}
\end{align}

% The resulting primal-dual algorithm using threshold dynamics (PD-TD) is summarized  in Algorithm \ref{alg:PD-TD}. Each update step is very simple and easy to implement.
We list the details of the proposed primal-dual algorithm using threshold dynamics (PD-TD) in Alg. \ref{alg:PD-TD}. 
\begin{algorithm}[t!]
\caption{Primal-dual algorithm using threshold dynamics (PD-TD)}\label{alg:PD-TD}
\begin{algorithmic}
\Require the region force term $f(x)$
\State choose parameters $\lambda$ and the maximum number of iterations $I$
\State initialize $u$ and $p$ to get $u^0(x)$ and $p^0$
\For{$k$ from 0 to $I-1$}
\State $\phi^k=\lambda f-(p^k)^2/4+p^kG_\sigma*(1-2u^k).$
\State $u^{k+1}= \mathbbm{1}_{\phi^k<0}$
\State $p^{k+1}=2 \langle u^{k+1},G_\sigma*(1-u^{k+1})\rangle/\langle u^{k+1},\mathbbm{1}_{\Omega}\rangle$
\If{stopping criterion is met}
\State stop iterations
\EndIf
\EndFor
\State return $u^{k+1}$
\end{algorithmic}
\end{algorithm}

\subsection{Primal-dual algorithm using soft threshold dynamics (PD-STD)}
% The PD-TD algorithm introduced before is very simple and efficient. 
% However, the binary constraints may affect its convergence sometimes. 
In this section, we propose another new primal-dual-based algorithm which is similar to the introduced PD-TD algorithm (Alg. \ref{alg:PD-TD}), including two steps of updating $u$ and $p$ at each iteration. But different from the PD-TD algorithm of Alg. \ref{alg:PD-TD}, we apply soft threshholding to $\phi^k$, instead of its hard threshholding as \eqref{hth}. This allows the result to be within $[0,1]$ and the whole procedure is differentiable and can thus be integrated into the DNNs as a variational sigmoid layer. 

% Instead of using the threshold dynamics, we can relax the original model (\ref{eq:primal2}) and apply the soft threshold dynamics. The soft threshold dynamic allows the model outputs to be within $[0,1]$. Besides, the whole procedure is differentiable and thus can be integrated into DNNs as a variational sigmoid layer, which will be discussed in the next subsection.

Given $u^k,p^k$ at the $(k+1)$-th iteration, we first fix $p^k$ and compute $u^{k+1}$ by
\begin{align}
    u^{k+1}=\arg\min_{u\in \mathcal{K}} \left\{L(u,p^k)+\epsilon\langle u,\log(u)\rangle + \epsilon\langle (1-u),\log(1-u)\rangle \right\}, \label{eq:primal_update_STD}
\end{align}
where an entropy regularization $\epsilon(\langle u,\log(u)\rangle + \langle 1-u,\log(1-u)\rangle)$ is introduced to the loss function $L(u, p^k)$ for some $\epsilon > 0$ \cite{liu2022deep}.

When $\epsilon \to 0^+$, it has been proved \cite{wang2017efficient} that the solution to \eqref{eq:primal_update_STD} becomes binary. 
% if $L$ is linear with respect to $u$. Therefore, after linearizing both equation (\ref{eq:primal_update_STD}) and equation (\ref{eq:primal_update_TD}), they become equivalent.

Second, we fix $u^{k+1}$ and introduce a proximal-point algorithm to update $p$:
\begin{align}
    p^{k+1}&=\arg\max_{p\in\mathbb{R}} \left\{L(u^{k+1},p)-\frac{1}{2\tau}|p-p^k|^2\right\}  \notag\\
    &=\frac{p^k+\tau\langle u^{k+1},G_\sigma*(1-u^{k+1})\rangle}{1+\tau\langle u^{k+1}.\mathbbm{1}_{\Omega}\rangle/2}. \label{eq:dual_update_STD}
\end{align}
The update (\ref{eq:dual_update_STD}) becomes (\ref{eq:dual_update_TD}) if we choose $\tau=+\infty$. After getting $u^{k+1}\in\mathcal{K}$, we can obtain a binary solution in $\mathcal{S}$ by thresholding $u^{k+1}$.
The details of the primal-dual algorithm using soft threshold dynamics (PD-STD) are listed in Alg. \ref{alg:PD-STD}. 
Notice that the PD-TD algorithm (Alg. \ref{alg:PD-TD}) can be taken as a special case of the PD-STD algorithm (Alg. \ref{alg:PD-STD}) when $\epsilon \to 0^+ $ and $\tau \to +\infty$.
\begin{algorithm}
\caption{Primal-dual algorithm using soft threshold dynamics (PD-STD)}\label{alg:PD-STD}
\begin{algorithmic}
\Require the region force term $f(x)$
\State choose parameters $\lambda$, $\epsilon$, $\tau$ and the maximum number of iterations $I$
\State initialize $u$ and $p$ to get $u^0(x)$ and $p^0$
\For{$k$ from 0 to $I-1$}
\State $\phi^k=\lambda f-(p^k)^2/4+p^kG_\sigma*(1-2u^k).$
\State $u^{k+1}= \exp(-\phi^k/\epsilon)/(1+\exp(-\phi^k/\epsilon))$
\State $p^{k+1}=(p^k+\tau\langle u^{k+1},G_\sigma*(1-u^{k+1})\rangle)/(1+\tau\langle u^{k+1},\mathbbm{1}_{\Omega}\rangle/2)$
\If{stopping criterion is met}
\State stop iterations
\EndIf
\EndFor
\State return $\mathbbm{1}_{u^{k+1}>0.5}$
\end{algorithmic}
\end{algorithm}

% In this section, we extend the PD-TD algorithm to the PD-STD algorithm by incorporating the soft threshold dynamics and propose a PD-STD sigmoid block that replaces the conventional sigmoid function in CNNs. Through this integration, we aim to incorporate the compactness prior into data-driven CNN models by considering the relaxed version of compactness model and the variational explanation for sigmoid..

\subsection{New PD-STD Neural Network\label{sec:dnns}}

% The task of integrating spatial priors, such as piecewise constant regions and shapes, into CNNs for semantic segmentation poses a challenge because existing techniques from variational segmentation models, including volume preservation, shape priors, and spatial regularization, cannot be directly employed due to the distinction between these two approaches.  Furthermore, the non-smoothness of cost functionals in variational models poses a risk of gradient explosion during backpropagation in DCNNs.  

% To address these challenges, the authors focus on the sigmoid function,
% \begin{align}
%     S(o)(x) = \frac{1}{1+e^{-o(x)}},
% \end{align}
% where $o: \Omega \to [0,1]$ is the logits output of the neural networks.  It acts as the binary classification function to enforce output probabilities before using the cross-entropy loss function. They proposed a variational viewpoint for the sigmoid activation function in \cite{LiuConvex}, enabling the integration of variational image segmentation techniques into DCNNs.  In the proposed framework, the sigmoid function can be seen as a dual function of smoothed ReLU.  

% Define the set $\mathcal{K}$ as:
% $$\mathcal{K}=\left\{u(x)\in[0,1],\forall x\in\Omega\bigg|u\in L^1(\Omega)\right\}.$$It is shown in \cite{LiuConvex} that the sigmoid operator $S$ is a minimizer of the following function:
% \begin{align}
%     \label{eq: variational explanation}
%     \min_{u \in \mathcal{K}} \left\{ <-o, u> + \epsilon <u, \log (u)>  \epsilon <1-u, \log (1-u)> \right\},
% \end{align}
% when $\epsilon = 1$.  

Utilizing the variational explanation of the sigmoid function \eqref{eq: variational explanation} and the neural network incorporating spatial regularization \eqref{eq: DNNwithRegularization}, we introduce a novel neural network which properly integrates the shape-compactness prior as follows:
\begin{equation}
\begin{cases} \label{eq:DNNwithCpt}
    {o}^t = \mathcal{T}_{{\Theta}^{t-1}}({v}^{t-1}, {v}^{t-2}, \dots, {v}^{0}),\  t=1, 2, \dots, T,\\
    {v}^{t} = \mathcal{A}^t({o}^{t}), t=1, 2, \dots, T-1,\\
    {v}^T = \arg\min_{u\in \mathcal{K}} \left\{\lambda\langle -{o}^T,u\rangle+\epsilon\langle u,\log(u)\rangle + \epsilon\langle (1-u),\log(1-u)\rangle +\frac{\langle u,G_\sigma*(1-u)\rangle^2}{\langle u,\mathbbm{1}_{\Omega}\rangle} \right\}.
\end{cases}
\end{equation}
In the last layer of the neural network above, we set the regularizer $\mathcal{R}(u)$ to the approximated  shape compactness prior  as defined  in (\ref{eq:primal2}). 
%Previous work integrated Total Variation (TV) \cite{jia2019regularized} into the activation function layer but required multiple iterations to calculate the dual variable. To overcome this, the soft threshold dynamics method, known for its faster and more stable dynamics compared to other TV regularization algorithms, was used in CNNs \cite{liu2022deep}.  In our paper, we want to integrate the compactness regularization term into the energy function to help CNNs generate more compact segmentation results.

The proposed PD-STD algorithm helps obtain smooth and compact segmentation regions, by unrolling the PD-STD algorithm as network sublayers: for the last layer $v^T$, each iteration of the proposed PD-STD algorithm can be regarded as a layer in DNNs, and thus the algorithm forms the new PD-STD Layers as shown in Fig. \ref{alg:PD-TD}, and this will be used as the last block for our neural network as shown in Fig. \ref{fig:architecture}. 
In Fig. \ref{fig:flowchart}, we show the layers which are resulted from Algorithm \ref{alg:PD-STD} and only $p$ and $u$ updates are illustrated. The $\phi$ updating is part of the $u$ updating and not shown here.  
% Replacing the original classification function sigmoid with model (\ref{eq:relaxedCompactness}) leads to the following network formulation. 
% \begin{align}
% f^{t} &= \mathcal{T}_{\Theta^{t-1}}(v^{t-1}), \notag\\
% v^t &= \mathcal{A}^t(f^t),
% \end{align}

% \begin{equation}
% \begin{cases}
%     \boldsymbol{o}^t = \mathcal{T}_{\boldsymbol{\Theta}^{t-1}}(\boldsymbol{v}^{t-1}, \boldsymbol{v}^{t-2}, \dots, \boldsymbol{v}^{0}),\  t=1, 2, \dots, T,\\
%     \boldsymbol{v}^{t} = \mathcal{A}^t(\boldsymbol{o}^{t}), t=1, 2, \dots, T-1,\\
%     \boldsymbol{v}^T = \arg\min_{u\in \mathcal{K}} \left\{\lambda\langle -\boldsymbol{o}^T,u\rangle+\epsilon\langle u,\log(u)\rangle +\frac{\langle u,G_\sigma*(1-u)\rangle^2}{\langle u,\mathbbm{1}_{\Omega}\rangle} \right\},
% \end{cases}
% \end{equation}

% where $\mathcal{T}_{\Theta^{t-1}}$ refers to the affine transformation used in the $(t-1)$-th layer of the neural network, and $\mathcal{A}^t$ denotes the activation function employed in the $t$-th layer. Furthermore, $v^{t}$ represents the output of the $t$-th layer of the neural network. 

%The PD-STD layers can help the DNNs generate more compact segmentation results.  
Our new  network architecture based on (\ref{eq:DNNwithCpt}) is shown in Fig. \ref{fig:architecture}.   The network in (\ref{eq:DNNwithCpt})   for $t=1,2,\cdots  T-1$ is referred as the backbone network. The backbone network can be any network which can produce the $o^T$ values for Algorithm \ref{alg:PD-STD}  which is used as the last layer for our new network.
%{\color{red} [Hao: what is last back?]}   
The backbone network  can be various image segmentation networks, including U-Net \cite{UNet} and SegNet \cite{SegNet}.  
%The last block with layers  with our proposed PD-STD sigmoid block while retaining all other network structures identical to a backbone network. 
%The choice of the backbone network can be determined mathematically by selecting different $\mathcal{T}_{{\Theta}}$. 

% Notably, the PD-STD Sigmoid block can be seamlessly integrated into CNN methods due to its plug-and-play nature. For example, we can incorporate the PD-STD block into a DeepLabV3 encoder-decoder structure. The details of the whole network are displayed in Figure \ref{fig:architecture}. The network architecture consists of a DeepLabV3 backbone, where the input is an image that undergoes processing. Subsequently, the output of the backbone passes through the Sigmoid activation function, which represents the DeepLabV3 method. When the output of the DeepLabV3 backbone is processed using the PD-STD function, it corresponds to the method proposed in our work.

\begin{figure}[t!]
    \centering
    \includegraphics[width=15cm]{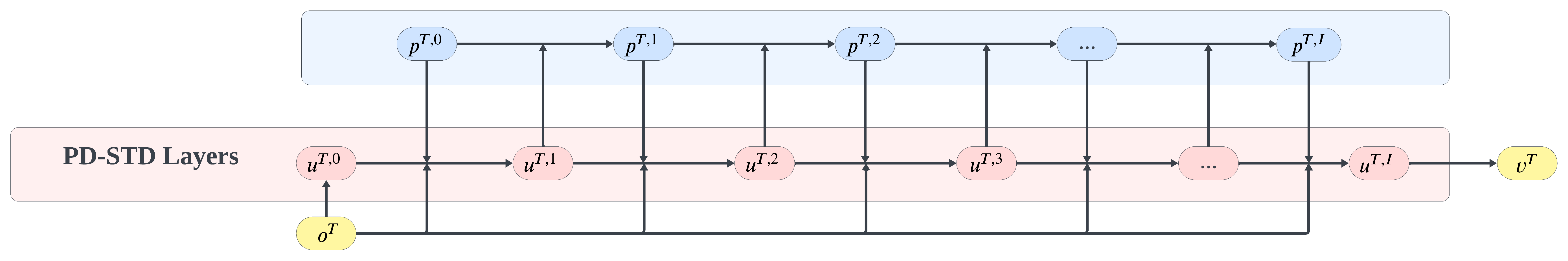}
    \caption{The network architecture of the proposed PD-STD Layers used for our proposed network as shown in Fig. \ref{fig:architecture}.  The red rectangle denotes the dual variables in the regularization space.  The blue rectangle denotes the dual variables in a compact-shaped space.}
    \label{fig:flowchart}
\end{figure}

\begin{figure}[t!]
    \centering
    \includegraphics[width=12cm]{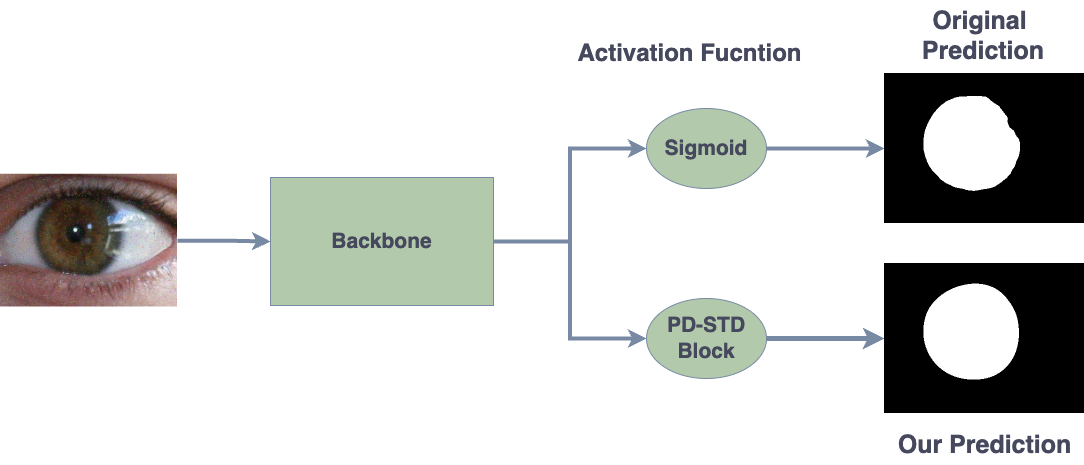}
    \caption{The architecture of a segmentation network with PD-STD layer.}
    \label{fig:architecture}
\end{figure}
% \subsection{Numerical implementation}
% The numerical implementation is very simple for our proposed algorithms, since there is no differential operators involved in the functional. The convolution is performed as the normal 2-dimensional discrete convolution with 0 padding. The discrete Gaussian kernel $G_\sigma$ is a matrix with specified size. Usually, we would choose the size to be an odd number $2n+1(n\in\mathbb{Z}^+)$. Then, the kernel $G_\sigma$ can be computed by the following Matlab function:
% \begin{lstlisting}
% function z=generateGaussianKernel(sigma, n):
%     [x,y]=meshgrid(-n:n,-n:n)
%     z=exp(-(x.^2+y.^2)/(2*sigma^2));
%     z=z/sum(z(:));
% end
% \end{lstlisting}
% The operations of convolutions with Gaussian kernels in Algorithm \ref{alg:PD-TD} and \ref{alg:PD-STD} can also be directly performed by the Matlab function: 
% \begin{lstlisting}
%     imgaussfilt(1-u,sigma,'FilterSize',2*n+1,'Padding',0);
% \end{lstlisting}
% or
% \begin{lstlisting}
%     imgaussfilt(1-2*u,sigma,'FilterSize',2*n+1,'Padding',0);
% \end{lstlisting}

\section{Numerical experiments} \label{sec: Numerical experiments}

In this section, experiments on both synthetic and real-world images are conducted to evaluate the performance of our proposed algorithms which are implemented in Matlab and Python, particularly the algorithms of PD-TD and PD-STD in Matlab and PD-STD-based DNNs in Pytorch. 
Implementation of the proposed PD-TD and PD-STD algorithms is straightforward as in Alg. \ref{alg:PD-TD} and Alg. \ref{alg:PD-STD}.
The two popular neural networks of DeepLabV3 \cite{DeeplabV3} and IrisParseNet \cite{IrisParseNet} are taken as the backbone networks to apply the introduced PD-STD block to build up the new PD-STD-based DNNs as stated in Section \ref{sec:dnns}. We compare the new networks with their original networks over different datasets to demonstrate the advantages of the PD-STD block, especially on noisy image datasets.
Training is conducted on a computer equipped with 4 x NVIDIA Tesla V100-32G GPUs. 
The convolution is performed using the classical 2-D discrete Gaussian kernel $G_\sigma$ of size $2n+1$ (with $n\in\mathbb{Z}^+$), with zero padding.
In this work, the grayvalue of each image pixel is normalized between 0 and 1. Two types of noise, i.e. Gaussian and salt-and-peppr, are applied for related experiments. 
Hence, adding Gaussian noise of $0.1$ to an image implies introducing noise with a standard deviation (SD) of $\rho = 0.1$ to each pixel.
The noise level for salt-and-pepper noise represents the probability of noise presence in the image.

We also compare the proposed algorithms with the state-of-the-art algorithms in efficiency, accuracy, and robustness. 
For the accuracy evaluation of image segmentation, the two metrics of Dice and IoU are applied:
% to measure the ratio of the intersection area and the union area of the ground-truth $\Omega_0$ and the computed segmentation $\Omega_r$:
\[
\text{Dice} = \frac{2 \times |\Omega_0 \cap \Omega_r| }{ |\Omega_0| + |\Omega_r|}\,, \quad \quad
\text{IoU} = \frac{{{|\Omega_0 \cap \Omega_r|}}}{{{|\Omega_0 \cup \Omega_r|}}},
\] 
where $\Omega_0$ is the ground truth label, $\Omega_r$ is the computed segmentation area, 
and ${|\Omega_0 \cap \Omega_r|}$ and ${|\Omega_0 \cup \Omega_r|}$ are their intersection and union, respectively.
% the area of intersection and union of the ground truth label $\Omega_0$ and our predicted segmentation result $\Omega_r$, respectively.
The following compactness metric is introduced to measure the shape compactness of the segmentation result $\Omega_r$:
\begin{equation}
\text{Compactness} = \frac{| \partial \Omega_r|^2}{|\Omega_r|} = \frac{(|u|_{TV})^2}{\int_{\Omega}u(x)dx}.
\end{equation}

% The dice coefficient and compactness are used as metrics for evaluating the performance of algorithms. The dice coefficient measures the accuracy of predictions, while the compactness metric assesses circularity.  The dice coefficient is defined as:
% \[
% \text{Dice} = \frac{2 \times |\Omega_0 \cap \Omega_r| }{ |\Omega_0| + |\Omega_r|},
% \]
% where $\Omega_0$ and $\Omega_r$ represent the ground truth label and our predicted segmentation result, respectively.  ${|\Omega_0 \cap \Omega_r|}$ and ${|\Omega_0 \cup \Omega_r|}$ are the area of intersection and union of $\Omega_0$ and $\Omega_r$, respectively.
% The compactness metric is evaluated by the shape compactness defined by equation (\ref{shapecptmetric}), as below:
% \begin{equation}
% \text{Compactness} = \frac{| \partial \Omega_r|^2}{|\Omega_r|} = \frac{(|u|_{TV})^2}{\int_{\Omega}u(x)dx}.
% \end{equation}

% The IoU metric also serves to measure the accuracy of predictions and is defined as follows:
% \[
% \text{IoU} = \frac{{{|\Omega_0 \cap \Omega_r|}}}{{{|\Omega_0 \cup \Omega_r|}}},
% \] 
% where ${|\Omega_0 \cap \Omega_r|}$ and ${|\Omega_0 \cup \Omega_r|}$ are the area of intersection and union of the ground truth label $\Omega_0$ and our predicted segmentation result $\Omega_r$, respectively.

\subsection{Experiments of the proposed PD-TD and PD-STD algorithms}
The experiment results of the proposed Algorithms of PD-TD and PD-STD, on both syntehtic and real-world images, are shown in this section and compared to the ADMM algorithm introduced by Dolz et al. \cite{dolz2017unbiased} (see Sec. \ref{sec:admm} for implementation details).
% the proposed Algorithms of PD-TD and PD-STD  \ref{alg:PD-STD} and \ref{alg:PD-TD} with the  ADMM algorithm of \cite{dolz2017unbiased} outlined in Section \ref{sec:admm} on both synthetic and real images to evaluate their performance. 
As shown in Fig. \ref{fig:syn}, 
 the segmentation result tends to be more circular as the weight parameter $\lambda$ in \eqref{eq:primal1} becomes smaller, which means the weight of the introduced shape-compactness regularization $C(u)$ is relatively bigger. 
 This is consistent with Lemma \ref{lemma:convergence}, and confirms that the introduced shape-compactness regularization $C(u)$ does work properly for all the three algorithms of ADMM, PD-TD and PD-STD.
% we observe that the segmentation results become more compact as the value of $\lambda$ decreases. This finding is consistent with Lemma \ref{lemma:convergence}. 
Additionally, both the PD-TD and PD-STD algorithms demonstrate the ability to produce more compact segmentation results, with the PD-STD algorithm achieving the best performance.

Fig. \ref{fig:real} illustrates the comparison on real images. Clearly, our proposed algorithms yield more compact segmentation results with smoother boundaries. To quantify the performance, we compare the averaged dice score, shape compactness, and computational time in Tab. \ref{table:real}. The results clearly indicate the superiority of our proposed methods: the proposed algorithms of PD-TD and PD-STD significantly outperform the ADMM algorithm in both compactness and efficiency. The compared ADMM algorithm even fails to obtain a compact segmentation region as result in some cases. In particular, the PD-STD algorithm  outperforms the other algorithms by producing results with the highest dice scores, which directly indicate more accurate segmentation results. Additionally, the PD-STD algorithm exhibits significantly faster computational speed compared to the ADMM algorithm. While the PD-TD algorithm demonstrates the shortest computational time, its compactness performance does not match the competitive level achieved by the PD-STD algorithm.

\begin{figure}[t!]
    \centering
    \includegraphics[width=12cm]{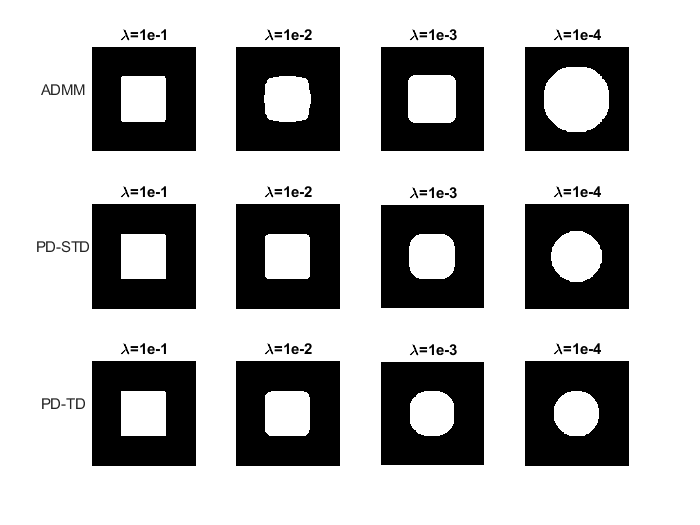}
    \caption{Experimental results of different algorithms with different $\lambda$ for comparison: it is obvious that the segmentation region tends to be more circular as the weight parameter $\lambda$ in \eqref{eq:primal1} becomes smaller (which means the weight of the introduced shape-compactness regularization $C(u)$ is bigger). This shows that the introduced shape-compactness regularization $C(u)$ does work properly for all the three algorithms of ADMM, PD-TD and PD-STD.}
    \label{fig:syn}
\end{figure}    

\begin{figure}[h!]
    \centering
    \includegraphics[width=13.8cm]{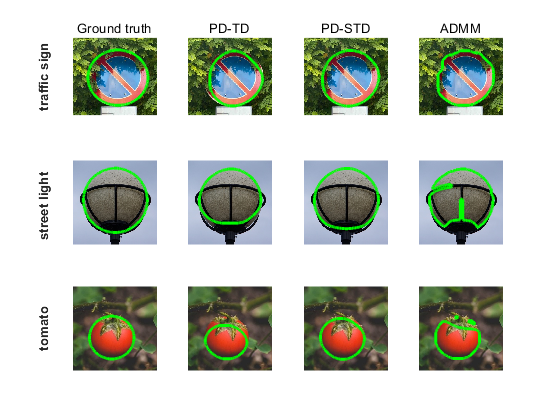}
    \caption{Segmentation results by different algorithms on some real images. The proposed algorithms of PD-TD and PD-STD generate more compact region with smoother boundaries as segmentation results; in contrast, the compared ADMM algorithm sometimes fails to obtain a compact segmentation region as result.}
    \label{fig:real}
\end{figure}

% \begin{table}[]
% \centering
% \caption{Averaged statistics of the PD-STD and ADMM algorithms}\label{table:real}
% \begin{tabular}{@{}cccc@{}}
% \toprule
%                 & \textbf{dice} & \textbf{compactness} & \textbf{time(s)} \\ \hline
% \textbf{PD-STD} & 0.9631              & 14.0086              & 6.8504                         \\
% \textbf{ADMM}   & 0.9246              & 22.2291              & 41.9574                        \\ \bottomrule
% \end{tabular}
% \end{table}

\begin{table}[t!]
\centering
\begin{tabular}{@{}cccc@{}}
\toprule
                & \textbf{Dice} & \textbf{Compactness} & \textbf{Time(s)} \\ \hline
\textbf{PD-TD} & 0.8863               & 14.0240              & 0.3936                         \\
\textbf{PD-STD} & 0.9631              & 14.0086              & 4.2269                         \\
\textbf{ADMM}   & 0.9246              & 22.2291              & 35.7028                        \\ \bottomrule
\end{tabular}
\caption{Averaged metrics of dice, compactness and computation time (s) are computed for performance comparison of PD-TD, PD-STD, and ADMM algorithms on real images shown in Fig. \ref{fig:real}.}\label{table:real}
\end{table}

% \subsection{PD-STD combined with DCNNs}
% In this section, we will evaluate the proposed method on noisy images and compare our method with the original DeepLabV3 network on several datasets.

\subsection{Experiments for PD-STD-Based DNNs}
% Based on the variational explanation of softmax layer, we can regard the PD-STD algorithm as an activation function PD-STD block and thus it can be applied on various CNNs.  
In this work, two popular DNNs of DeepLabV3 \cite{DeeplabV3} and IrisParseNet \cite{IrisParseNet} are taken as the backbone networks for which the PD-STD block is introduced to replace their sigmoid layers as shown in Fig. \ref{fig:architecture}, which encodes the compact shape prior in the proposed DNNs for training.
Experiments show great performance of the proposed PD-STD-based DNNs in extracting compact regions from images, especially when there is high noise.
% we take the  DeepLabV3\cite{DeeplabV3} as the backbone network and apply our PD-STD block to it to evaluate the performance of our method given in Section \ref{sec:dnns}.  Our approach is also compared with the powerful network IrisParseNet\cite{IrisParseNet} on a dataset and shows a leading performance on various benchmarks especially when we add relatively high level noise to the testing images.

\subsubsection{Implementation Details}
%\paragraph{}\  
% Our proposed method (c.f. section \ref{sec:dnns}) is implemented under pytorch on a computer equipped with 4 x NVIDIA Tesla V100-32G GPUs.  
% To accelerate the training, we import 'deeplabv3\_resnet50' from 'torchvision.models. segmentation' and set the 'pretrained\_backbone=TRUE' to load the pre-training parameters for both DeepLabV3 and our method.  
% Similarly, the pre-training weights for IriParseNet and the modified IrisParseNet by adding our PD-STD block are the same and it can be found on [url]. 
% As for the IrisParseNet, the original code is written in caffe and it can be downloaded from the official implementation https://github.com/xiamenwcy/IrisParseNet.  
We reimplemented the networks of DeepLabV3 and IrisParseNet as described in \cite{DeeplabV3} and \cite{IrisParseNet} using PyTorch and strictly adhered to the hyper-parameter settings specified in the papers \cite{DeeplabV3} and \cite{IrisParseNet}. We conducted multiple runs by randomly initializing the network parameters and selected the best results from each run for both networks.
'deeplabv3\_resnet50' from 'torchvision.models. segmentation' is imported and we set the 'pretrained\_backbone=TRUE' to load the pre-training parameters for all the networks, so as to accelerate training. 
In our experiments, we use the ADAM optimizer and the learning rate is set as 0.0001. 
Actually, both the weight parameter $\epsilon$ in \eqref{eq:primal_update_STD} and the iteration number $I$ in Alg. \ref{alg:PD-STD} impact the segmentation results.
Increasing the iteration number $I$ enhances shape-compactness of the segmentation results, albeit at the cost of more running time. On the other hand, $\epsilon$ influences the speed at which the segmentation result reaches optimum: Higher values of $\epsilon$ often speed up the optimization process. Consequently, we aim to select an optimal pair of $\epsilon$ and $I$ to achieve rapid convergence and shape compactness jointly in our experiments.  
% For fair comparisonm   of PD-STD on noisy images and compare the performance of our method with the popular DeepLabV3+ network and the state-of-the-art IrisParseNet network.  

% For fair comparison, we use the same parameters to train the DeepLabV3, IrisParseNet and the modified networks and replace the Sigmoid layer in these two networks by our PD-STD  block as shown in Figure \ref{fig:architecture}. 
% The constant  $\epsilon$ is a weight parameter in equation (\ref{eq:primal_update_STD}), while $I$ represents the iteration number in Algorithm \ref{alg:PD-STD}. Both $\epsilon$ and $I$ significantly impact the circularity of the segmentation results. By applying the PD-STD algorithm, we can generate more compact segmentation results. Increasing the iteration number enhances the compactness of the segmentation results, albeit at the cost of increased running time. On the other hand, $\epsilon$ influences the speed at which the segmentation results attain circularity. Higher values of $\epsilon$ expedite the circularization process. Consequently, we aim to select an optimal pair of $\epsilon$ and iteration number $I$ to achieve both rapid convergence and compactness in the results.  

\subsubsection{Experiments on Iris Dataset}
%\paragraph{}\  

% This section aims to assess the performance of our network with the PD-STD  block on the Iris dataset, which consists of 483 training images and 436 testing images, all of size $400 \times 300$. 
In this section, the Iris dataset of UBIRIS.v2, consisting of 483 training images and 436 testing images (all of size $400 \times 300$), is taken for experiments to show the performance of our proposed PD-STD-based network. 
This dataset is publically available on the GitHub website: https://github.com/xiamenwcy/IrisParseNet.
% In real-world scenarios, images are often affected by noise during acquisition, compression, and transmission, so 
We add different degrees of Gaussian noise and salt-and-pepper noise to the test image dataset to further evaluate the robustness of our proposed networks.

The two networks of DeepLabV3 and our proposed PD-STD + DeepLabV3 are first trained by the clean Iris dataset (without noise), and the iteration number $I$ of the PD-STD block is set to $50$. 
% This study focuses on training the DeepLabV3 model and our proposed model, which  replaces the sigmoid layer in DeepLabV3 with our PD-STD  block, using a clean (no added noise)  training dataset.  
The two networks are then evaluated on both clean and noisy image datasets. 
Different levels of Gaussian noise with standard deviation ($0.1$, $0.07$, $0.05$, and $0.01$) and salt-and-pepper noise ($0.01$) are also applied for tests.
% To better evaluate the robustness, we add different levels of Gaussian noise with standard deviation ($0.1$, $0.07$, $0.05$, and $0.01$) and salt-and-pepper noise ($0.01$)  to the testing dataset when we apply the training model to the testing dataset. 
% For example, adding $0.1$ Gaussian noise to an image implies adding Gaussian noise with a standard deviation (SD) of $\rho = 0.1$ to each pixel. 
% The noise level for salt-and-pepper noise represents the probability of noise presence in the image.  
Experiment results for the proposed method versus the original DeepLabV3 network are shown in Tab. \ref{table:trainOnClean},
% To provide a comprehensive performance comparison, we present a detailed analysis of the proposed method versus the backbone DeepLabV3 network for iris segmentation in Table \ref{table:trainOnClean}. 
where higher values of IoU and Dice metrics indicate more accurate results and the metric of Compactness approaching $4\pi \approx 12.56 $ signifies more compact segmentation regions. 
Clearly, our proposed PD-STD + DeepLabV3 outperforms DeepLabV3 in all metrics, particularly when segmenting images with higher noise levels. Fig. \ref{fig:trainOnCleanDeepLabv3} shows several examples that illustrate the effectiveness of the proposed PD-STD + DeepLabV3 in mitigating the rough boundaries caused by the noise introduced. It is obvious that PD-STD + DeepLabV3 yields segmentation outcomes that are more preferable with compact shapes. When the Gaussian noise level is relatively large (such as 0.1), DeepLabV3 fails to get reasonable segmentation results.  

% The performance of the algorithms on the test set (456 images).    And the experiment result can be found in Table \ref{table:Iris}.  It is shown that our PD-STD algorithm can improve the mIoU and the dice about $1\%$ compared to the original DeepLabV3.  In Figure \ref{fig:iris}, we visualize the segmentation results for easy comparisons.  It is easy to see that the proposed method can fight against the noise and produce better segmentation.  

\begin{table}[t!]
\centering

\resizebox{\textwidth}{20mm}{
\begin{tabular}{@{}c|c|c|cccc|c@{}}
\hline
    & Method &Clean & \multicolumn{4}{c|}{Gaussian}  &  Salt \& Pepper \\ \hline
Noise level    &        &   0   &0.01 & 0.05 & 0.07 & 0.1 &     0.01   \\
                \hline
 \multirow{2}*{IoU}& DeepLabV3 & {0.9096} & {0.9089} & {0.8572}  & {0.7315}  & {0.3817}  & {0.8330}    \\
 ~ &PD-STD + DeepLabV3 & $\boldsymbol{0.9102}$  & $\boldsymbol{0.9096}$  &  $\boldsymbol{0.8651}$ & $\boldsymbol{0.7793}$ &$\boldsymbol{0.5511}$ & $\boldsymbol{0.8555}$    \\
\hline
  \multirow{2}*{Dice}& DeepLabV3 & {0.9439}  & {0.9435}  & {0.9086} &{0.7979}  & {0.4347}  & {0.8904}   \\
 ~ &PD-STD + DeepLabV3 & $\boldsymbol{0.9442}$  & $\boldsymbol{0.9439}$ & $\boldsymbol{0.9106}$ & $\boldsymbol{0.8389}$  & $\boldsymbol{0.6168}$  & $\boldsymbol{0.9055}$   \\
  \hline
  \multirow{2}*{Compactness}& DeepLabV3 & {14.1590}  & {14.1299} & {15.1426} & {18.7855}  & {19.4549}  & {16.4107}    \\
 ~ &PD-STD + DeepLabV3 & $\boldsymbol{13.7665}$  & $\boldsymbol{13.8243}$  &$\boldsymbol{14.3454}$ & $\boldsymbol{15.0243}$  & $\boldsymbol{13.2463}$  & $\boldsymbol{14.6427}$    \\
 \hline
\end{tabular}
}
\caption{Comparison btw. DeepLabV3 and our proposed PD-STD + DeepLavV3 when training with a clean image dataset. Our proposed method of PD-STD + DeepLabV3 obtains better results in all metrics of IoU, Dice and Compactness. Especially, when the Gaussian noise level reaches relatively (0.1), DeepLabV3 fails to get reasonable segmentation results, but our proposed PD-STD + DeepLabV3 still yields meaningful results. Please see Fig. \ref{fig:trainOnCleanDeepLabv3} for illustrations.} \label{table:trainOnClean}
\end{table}

\begin{figure*}[t!]
  \centering 
  \begin{minipage}[b]{\linewidth} 
  \subfloat[Noisy Image]{
    \begin{minipage}[b]{0.23\linewidth}
      \centering
      \includegraphics[width=\linewidth]{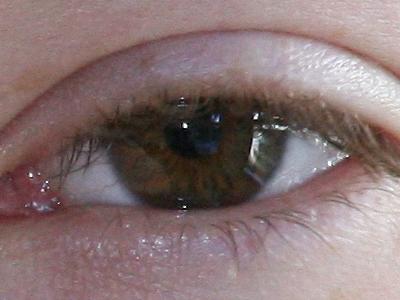}\vspace{5pt}
      \includegraphics[width=\linewidth]{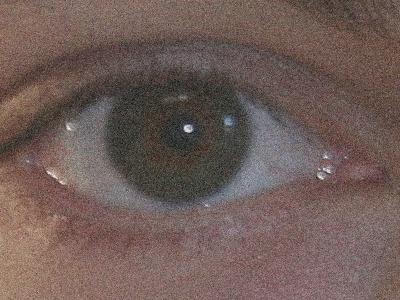}\vspace{5pt}
      \includegraphics[width=\linewidth]{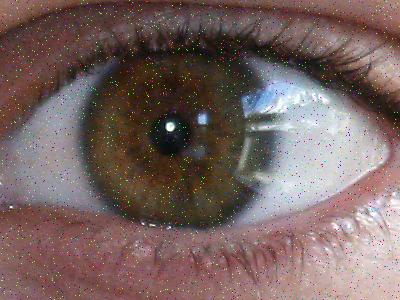}
    \end{minipage}
  }
  \hfill
   \subfloat[Ground Truth]{
    \begin{minipage}[b]{0.23\linewidth}
      \centering
      \includegraphics[width=\linewidth]{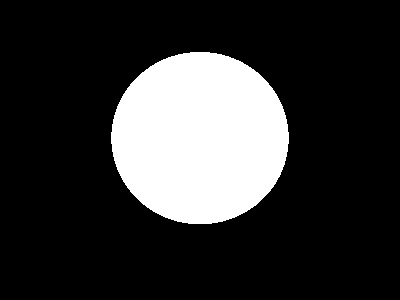}\vspace{5pt}
      \includegraphics[width=\linewidth]{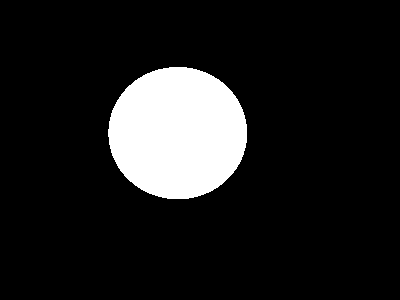}\vspace{5pt}
      \includegraphics[width=\linewidth]{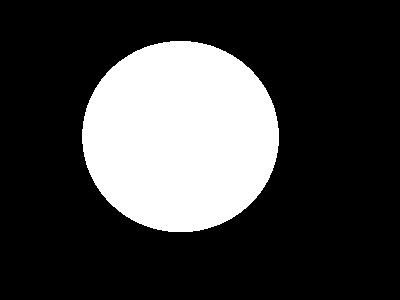}
    \end{minipage}
  }
  \hfill
    \subfloat[DeepLabV3]{
    \begin{minipage}[b]{0.23\linewidth}
      \centering
      \includegraphics[width=\linewidth]{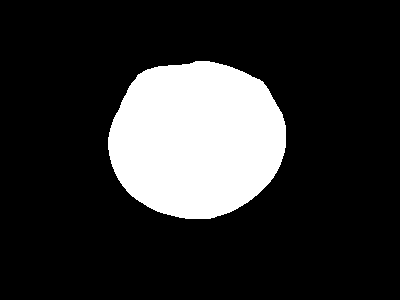}\vspace{5pt}
      \includegraphics[width=\linewidth]{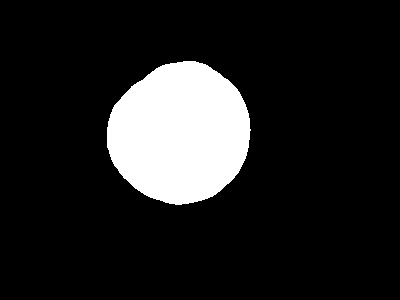}\vspace{5pt}
      \includegraphics[width=\linewidth]{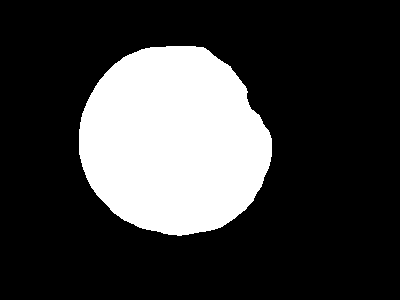}
    \end{minipage}
  }
  \hfill
    \subfloat[PD-STD + DeepLabV3]{
    \begin{minipage}[b]{0.23\linewidth}
      \centering
      \includegraphics[width=\linewidth]{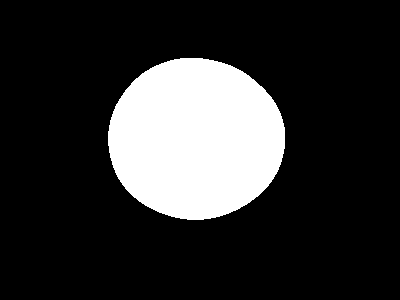}\vspace{5pt}
      \includegraphics[width=\linewidth]{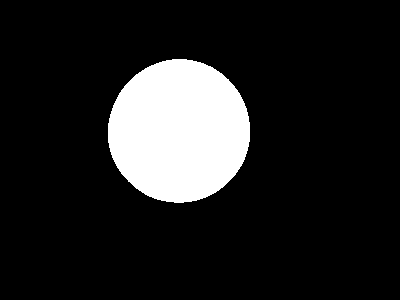}\vspace{5pt}
      \includegraphics[width=\linewidth]{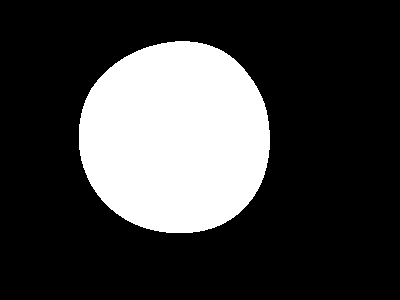}
    \end{minipage}
  }
  \end{minipage}
  \vfill
  \caption{Segmentation results predicted by DeepLabV3 and our methods of PD-STD + DeepLabV3 trained on a clean image dataset.  Noise level of the test image dataset from top to bottom: Gaussian noise level of $0.01$, $0.07$, salt and pepper noise level of $0.01$.}\label{fig:trainOnCleanDeepLabv3}
\end{figure*}

\begin{table}[t!]
\centering

\resizebox{\textwidth}{18mm}{
\begin{tabular}{@{}c|c|c|cccc|c@{}}
\hline
    & Method &Clean & \multicolumn{4}{c|}{Gaussian}  &  Salt \& Pepper \\ \hline
Noise level    &        &  0    &0.1  &   0.15 & 0.17 & 0.2 &   0.05     \\
                \hline
 \multirow{2}*{IoU}& IrisParseNet &$\boldsymbol{0.9054}$     & {0.8209}  & {0.4363} & {0.2769} & {0.1121}  & {0.5822} \\
 ~ &PD-STD + IrisParseNet & ${0.9030}$  & $\boldsymbol{0.8794}$ & $\boldsymbol{0.7987}$ & $\boldsymbol{0.7239}$  & $\boldsymbol{0.5836}$ & $\boldsymbol{0.7554}$   \\
 \hline
  \multirow{2}*{Dice}& IrisParseNet & $\boldsymbol{0.9415}$    & {0.8803}  & {0.5357} & {0.3682}  & {0.1649}   & {0.6817} \\
 ~ &PD-STD + IrisParseNet & ${0.9400}$   & $\boldsymbol{0.9249}$  & $\boldsymbol{0.8675}$ & $\boldsymbol{0.8070}$  & $\boldsymbol{0.6788}$    & $\boldsymbol{0.8341}$   \\
  \hline
  \multirow{2}*{Compactness}& IrisParseNet & {14.8976}    & {24.9740}  & {44.1741}  & {45.6170}  & {35.8653}   & {61.6093}\\
 ~ &PD-STD + IrisParseNet & $\boldsymbol{13.9134}$ & $\boldsymbol{13.9971}$ & $\boldsymbol{14.0876}$  & $\boldsymbol{14.3608}$ & $\boldsymbol{13.8360}$    & $\boldsymbol{13.9030}$   \\
 \hline
\end{tabular}
}
\caption{
%Comparison with IrisParseNet when training with a clean dataset.
Comparison of IrisParseNet and our proposed PD-STD + IrisParseNet when training with a clean image dataset. Our proposed method of PD-STD + IrisParseNet performs similarly as IrisParseNet when testing on clean images, but significantly outperforms IrisParseNet when testing on noisy images.
Especially, when the Gaussian noise level increases, IrisParseNet fails to get reasonable segmentation results, but our proposed PD-STD + IrisParseNet still works properly. Please see Fig. \ref{fig:trainOnCleanIrisParse} for illustrations.
} \label{table:trainOnCleanIrisParse}
\end{table}

\begin{figure*}[t!]
  \centering 
  \begin{minipage}[b]{\linewidth} 
  \subfloat[Noisy Image]{
    \begin{minipage}[b]{0.23\linewidth}
      \centering
      \includegraphics[width=\linewidth]{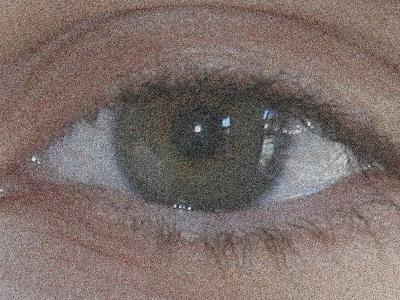}\vspace{5pt}
      \includegraphics[width=\linewidth]{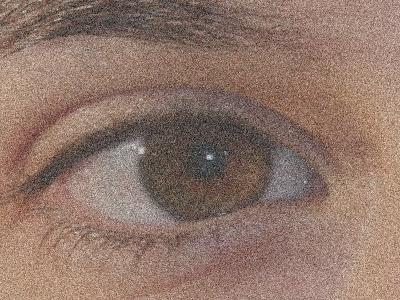}\vspace{5pt}
      \includegraphics[width=\linewidth]{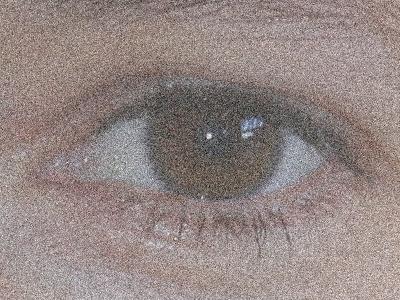}\vspace{5pt}
      \includegraphics[width=\linewidth]{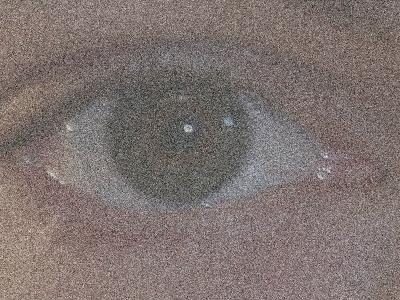}\vspace{5pt}
      \includegraphics[width=\linewidth]{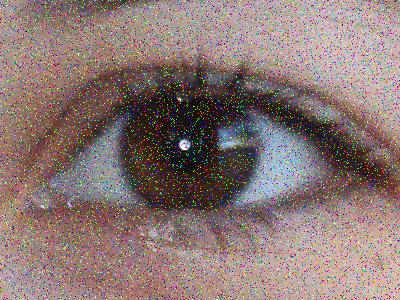}
    \end{minipage}
  }
  \hfill
   \subfloat[Ground Truth]{
    \begin{minipage}[b]{0.23\linewidth}
      \centering
      \includegraphics[width=\linewidth]{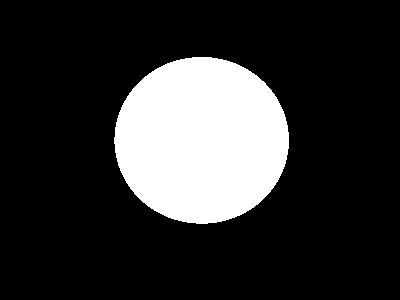}\vspace{5pt}
      \includegraphics[width=\linewidth]{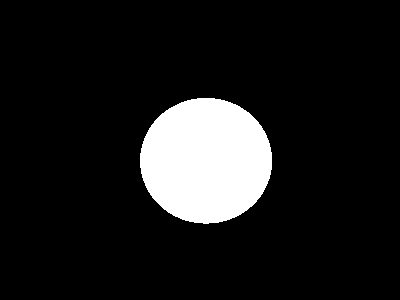}\vspace{5pt}
      \includegraphics[width=\linewidth]{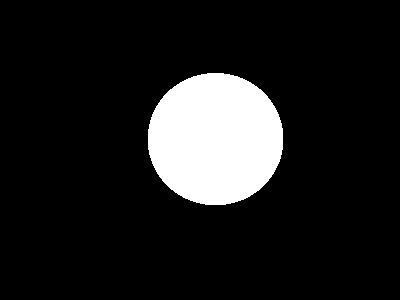}\vspace{5pt}
      \includegraphics[width=\linewidth]{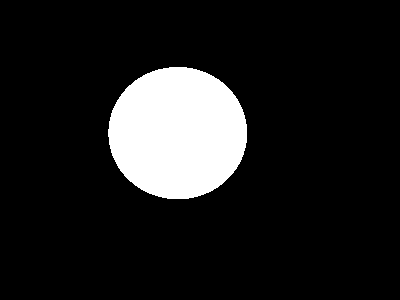}\vspace{5pt}
      \includegraphics[width=\linewidth]{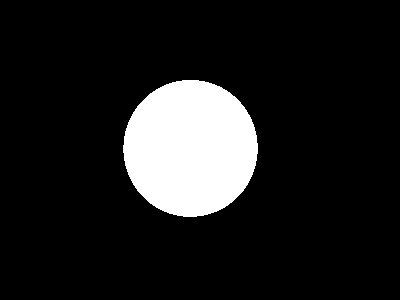}
    \end{minipage}
  }
  \hfill
    \subfloat[IrisParseNet]{
    \begin{minipage}[b]{0.216\linewidth}
      \centering
      \includegraphics[width=\linewidth]{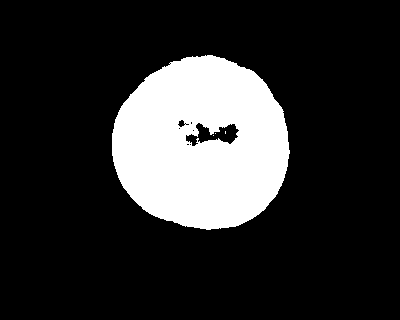}\vspace{5pt}
      \includegraphics[width=\linewidth]{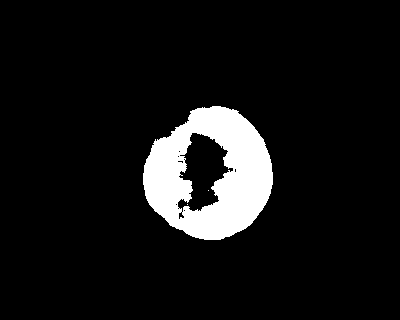}\vspace{5pt}
      \includegraphics[width=\linewidth]{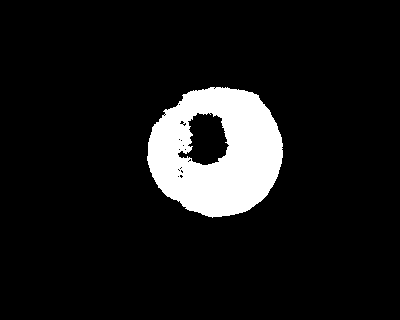}\vspace{5pt}
      \includegraphics[width=\linewidth]{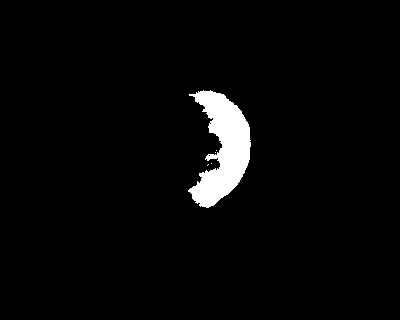}\vspace{5pt}
      \includegraphics[width=\linewidth]{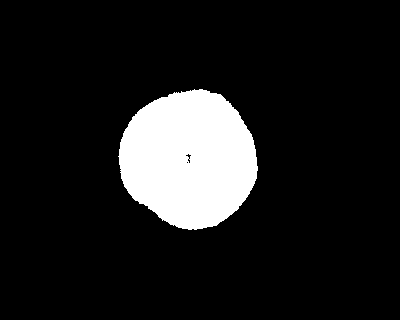}
    \end{minipage}
  }
  \hfill
    \subfloat[PD-STD+IrisParseNet]{
    \begin{minipage}[b]{0.216\linewidth}
      \centering
      \includegraphics[width=\linewidth]{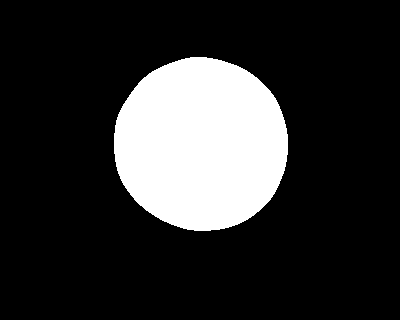}\vspace{5pt}
      \includegraphics[width=\linewidth]{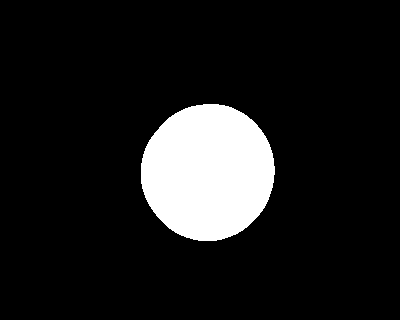}\vspace{5pt}
      \includegraphics[width=\linewidth]{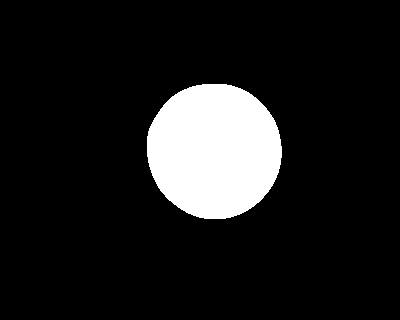}\vspace{5pt}
      \includegraphics[width=\linewidth]{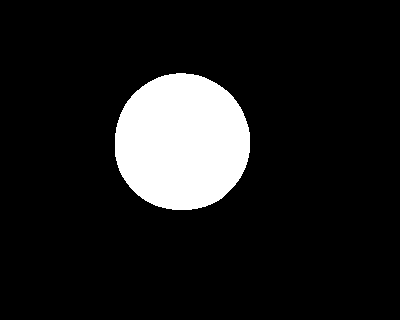}\vspace{5pt}
      \includegraphics[width=\linewidth]{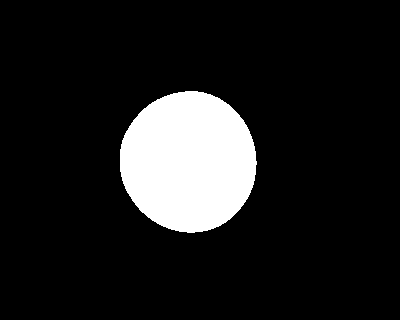}
    \end{minipage}
  }
  \end{minipage}
  \vfill
  \caption{Segmentation results predicted by IrisParseNet and our proposed PD-STD+IrisParseNet trained on clean image dataset.  Noise level from top to bottom: Gaussian noise level of $0.1$, $0.15$, $0.17$, $0.2$, salt-and-pepper noise level of $0.05$.
 It is visually obvious that our porposed PD-STD + IrisParseNet works much more robustly than the popular IrisParseNet in segmenting noisy images.} \label{fig:trainOnCleanIrisParse}
\end{figure*}

Now we introduce the proposed PD-STD block to the state-of-the-art IrisParseNet network \cite{IrisParseNet}, i.e. the PD-STD+IrisParseNet, to further show its effectiveness in pursuit of shape prior information on compactness.  
In contrast to the general-purpose DeepLabV3 network, IrisParseNet is purposely designed to tackle the challenge of iris images affected by severe noise.  
% In pursuit of this goal, we compare the performance of our network with the PD-STD block with that of the state-of-the-art IrisParseNet network \cite{IrisParseNet}.  
Similar experiments for IrisParseNet \cite{IrisParseNet} and our proposed PD-STD + IrisParseNet are conducted for comparisons.
Different levels of Gaussian noise ranging from 0.1 to 0.2 and salt-and-pepper noise of 0.05 are set for the experiments.  
Tab. \ref{table:trainOnCleanIrisParse} demonstrates that our proposed PD-STD + IrisParseNet exhibits comparable performance in segmenting clean images, and significantly outperforms the original IrisParseNet in segmenting noisy images, as the examples illustrated in Fig. \ref{fig:trainOnCleanIrisParse}. 
% This exactly approves that the introduced PD-STD block is able to enforce the compactness of the segmentation region, with a more single and compact region as the result.
% It can be observed that the compact regularization effects of the PD-STD  block enhance segmentation performance by filling in gaps and generating outputs that closely resemble the ground truth.

\begin{figure*}[t!]
  \centering 
  \begin{minipage}[b]{1.0\textwidth} 
  {
    \begin{minipage}[b]{0.17\textwidth}
      Noisy Image \   \  \vspace{55pt} \\
      Ground Truth\  \vspace{55pt} \\
      IrisParseNet\ \   \vspace{55pt} \\
      DeepLabV3\  \vspace{45pt} \\
      IrisParseNet + PD-STD \  \vspace{45pt} \\
      DeepLabV3 + PD-STD \   \\
    \end{minipage}
  }
  \hfill
     \subfloat[]{
    \begin{minipage}[b]{0.18\textwidth}
      \centering
      \includegraphics[width=\linewidth]{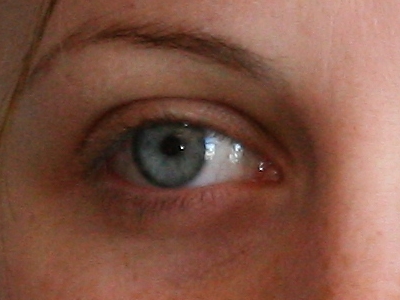}\vspace{5pt}
      \includegraphics[width=\linewidth]{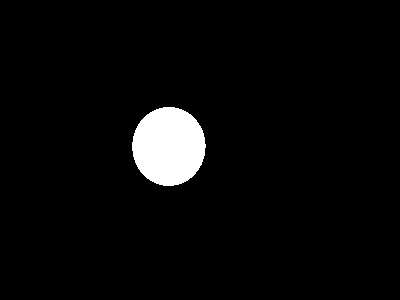}\vspace{5pt}
      \includegraphics[width=\linewidth]{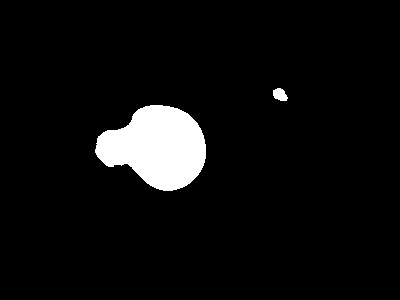}\vspace{5pt}
      \includegraphics[width=\linewidth]{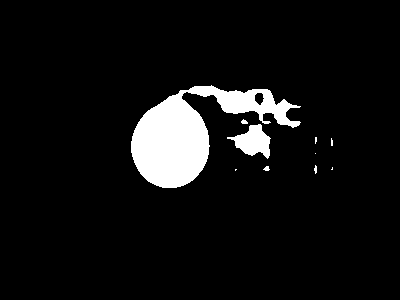}\vspace{5pt}
      \includegraphics[width=\linewidth]{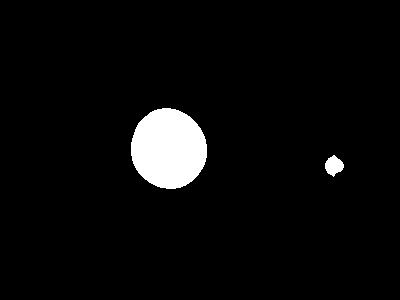}\vspace{5pt}
      \includegraphics[width=\linewidth]{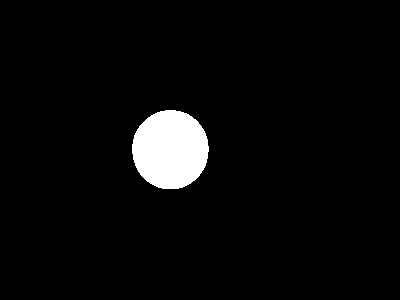}
    \end{minipage}
  }
  \hfill
    \subfloat[]{
    \begin{minipage}[b]{0.18\textwidth}
      \centering
      \includegraphics[width=\linewidth]{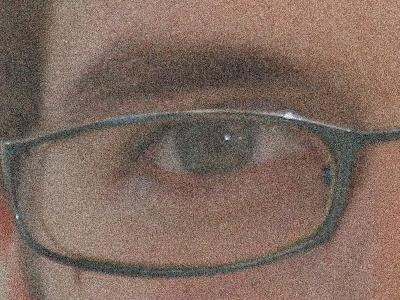}\vspace{5pt}
      \includegraphics[width=\linewidth]{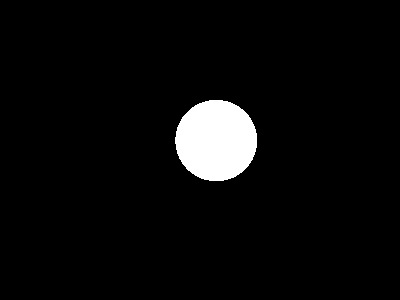}\vspace{5pt}
      \includegraphics[width=\linewidth]{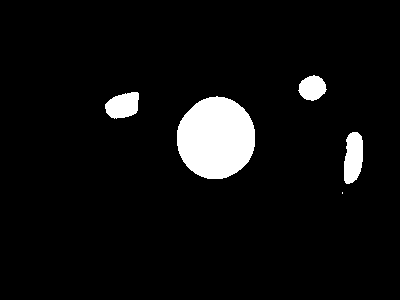}\vspace{5pt}
      \includegraphics[width=\linewidth]{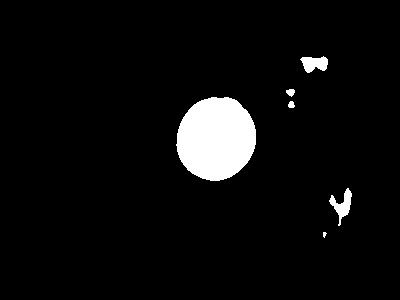}\vspace{5pt}
      \includegraphics[width=\linewidth]{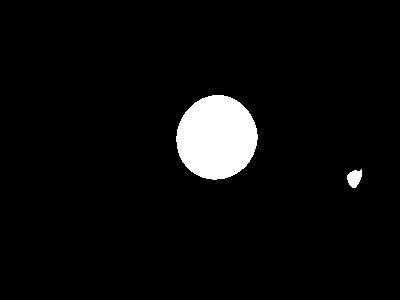}\vspace{5pt}
      \includegraphics[width=\linewidth]{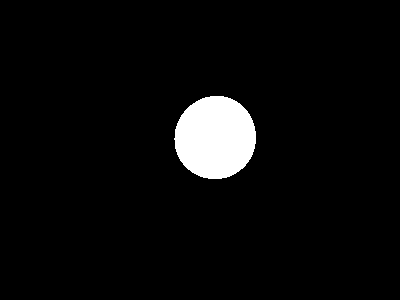}
    \end{minipage}
  }
  \hfill
   \subfloat[]{
    \begin{minipage}[b]{0.18\textwidth}
      \centering
      \includegraphics[width=\textwidth]{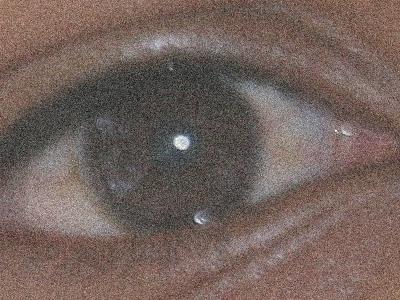}\vspace{5pt}
      \includegraphics[width=\linewidth]{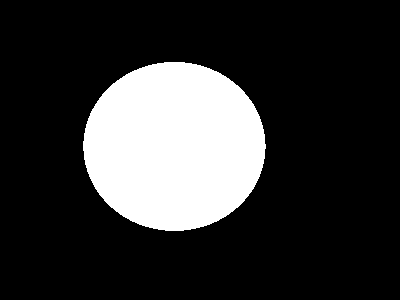}\vspace{5pt}
      \includegraphics[width=\linewidth]{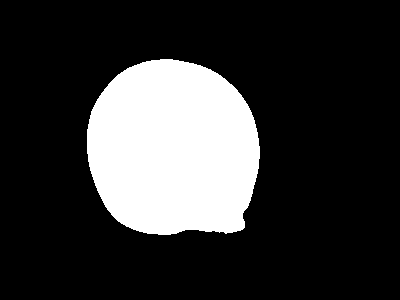}\vspace{5pt}
      \includegraphics[width=\linewidth]{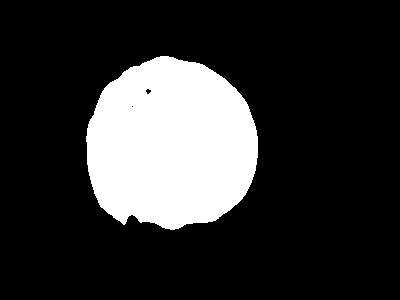}\vspace{5pt}
      \includegraphics[width=\linewidth]{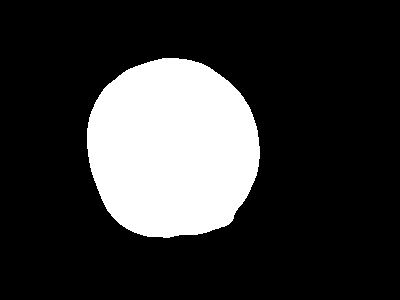}\vspace{5pt}
      \includegraphics[width=\linewidth]{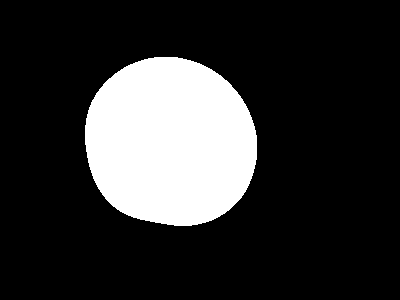}
    \end{minipage}
  }
  \hfill
    \subfloat[]{
    \begin{minipage}[b]{0.18\textwidth}
      \centering
      \includegraphics[width=\linewidth]{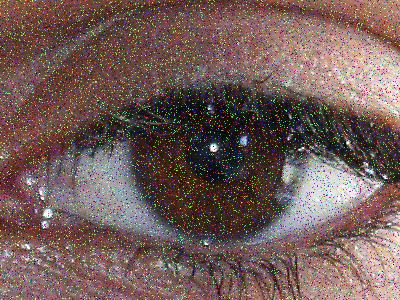}\vspace{5pt}
      \includegraphics[width=\linewidth]{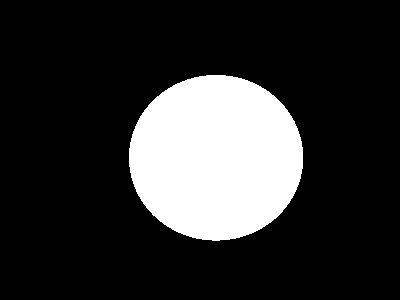}\vspace{5pt}
      \includegraphics[width=\linewidth]{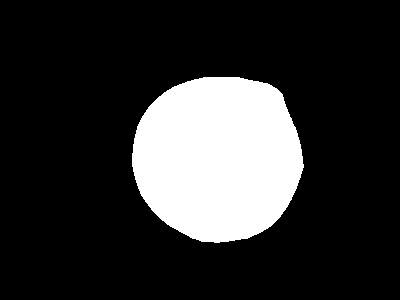}\vspace{5pt}
      \includegraphics[width=\linewidth]{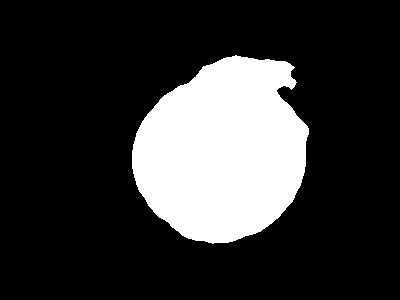}\vspace{5pt}
      \includegraphics[width=\linewidth]{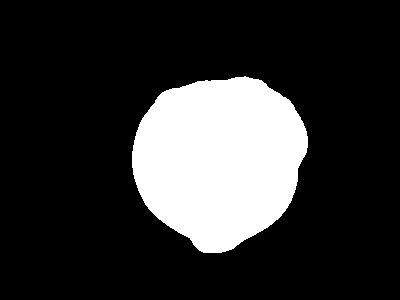}\vspace{5pt}
      \includegraphics[width=\linewidth]{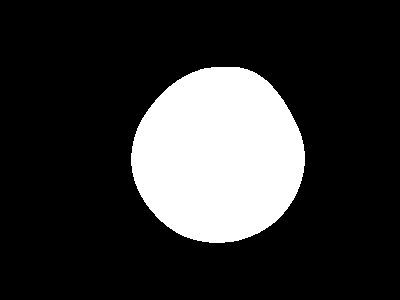}
    \end{minipage}
  }
  \end{minipage}
  \vfill
  \caption{
  Some illustrative examples are given in this figure: the four networks of DeepLabV3, IrisParseNet, PD-STD+DeepLabV3 and PD-STD+IrisParseNet are trained on the image dataset of noise level $0.1$, and their performance on the test image of noise level: no noise, Gaussian noise $0.05$ and $0.1$, Salt \& Pepper $0.05$, is shown in Column (a) - (d) respectively.
 The trained networks of DeepLabV3 and IrisParseNet work worse on the test images of no noise and Gaussian noise $0.05$; in contrast, the proposed networks of PD-STD+DeepLabV3 and PD-STD+IrisParseNet work reliably on the test images of various noise levels.
  % Segmentation results predicted by DeepLabV3 and our methods trained on the dataset with $0.1$ Gaussian noise.  Noise level for (a) - (d) are: no noise, Gaussian noise level of $0.05$, $0.1$, salt and pepper noise level of $0.05$.  
  }\label{fig:trainOnNoisyDeepLabv3}
\end{figure*}

On the other hand, we also build up a training image dataset with a Gaussian noise level of $0.1$, and compare the performance of the four networks of DeepLabV3, IrisParseNet and our proposed PD-STD-based networks, trained on the noisy image dataset with those trained on the clean image dataset.
% The inclusion of noisy images in the training dataset can enhance the robustness of the trained model. To assess this, we add Gaussian noise with a noise level of $0.1$ to each training image and compare the performance of models trained on a noisy dataset with those trained on a clean dataset. 
Tab. \ref{table:trainOnnoisy} presents the segmentation results on clean images and images with different noise levels, including Gaussian noise levels of $0.1$, $0.05$, and $0.01$, as well as salt-and-pepper noise levels of $0.05$ and $0.01$. As shown in Tab. \ref{table:trainOnnoisy} and Fig. \ref{fig:trainOnNoisyDeepLabv3}, 
the performance of DeepLabV3 and IrisParseNet is largely improved when trained in the noisy image dataset. However,
our proposed PD-STD-based networks still outperform in all cases, especially in the compactness metric when segmenting images with Salt \& Pepper noises. 
% our proposed PD-STD block is able to convergence to a better optimum and improve segmentation results.

\begin{table}[t!]
\centering

\resizebox{\textwidth}{30mm}{
\begin{tabular}{@{}c|c|c|ccc|cc@{}}
\hline
    & Method &Clean & \multicolumn{3}{c|}{Gaussian}  & \multicolumn{2}{c}{Salt \& Pepper} \\ \hline
Noise level    &        &    0  &0.01 & 0.05 & 0.1 & 0.01 & 0.05       \\
                \hline
 \multirow{4}*{IoU}& IrisParseNet & {0.8618}    &  {0.8943}   & {0.8995}  &  {0.9017} & {0.8629}  & {0.8491} \\
 ~ & DeepLabV3 & ${0.8642}$  &  ${0.8737}$ & ${0.8998}$  & ${0.8999}$  & ${0.8763}$  & ${0.6718}$  \\
 ~ &PD-STD + IrisParseNet & {0.8880}  & $\boldsymbol{0.9002}$ & ${0.9042}$  & $\boldsymbol{0.9044}$  & {0.8827}  & $\boldsymbol{0.8654}$\\
 ~ & PD-STD + DeepLabV3 & $\boldsymbol{0.8933}$  & {0.8959}  & $\boldsymbol{0.9051}$  & ${0.9030}$ & $\boldsymbol{0.8942}$  & {0.6784}  \\
 \hline
  \multirow{4}*{Dice}& IrisParseNet & {0.9097}    &  {0.9330}   & {0.9370}  & {0.9388}  & {0.9108}  &{0.9032}  \\
 ~ & DeepLabV3 & {0.9120}  & {0.9186}  & {0.9373}  & {0.9371}  &  {0.9215} & {0.7531}  \\
 ~ & PD-STD + IrisParseNet & {0.9293}  & $\boldsymbol{0.9371}$ & {0.9397}  & $\boldsymbol{0.9408}$  & {0.9263}  &  $\boldsymbol{0.9156}$\\
 ~ &PD-STD + DeepLabV3 & $\boldsymbol{0.9333}$  & {0.9350}  & $\boldsymbol{0.9411}$  & {0.9397}  &  $\boldsymbol{0.9339}$  & {0.7617} \\
  \hline
  \multirow{4}*{Compactness}& IrisParseNet & {19.1554}    &  {15.7192}   & {15.1328}  & {14.7285}  & {19.1122} &  {19.3267}  \\
 ~ &DeepLabV3 & {21.0197}  & {19.6417}  & {15.6534}  & {14.6295}  & {18.5644}  & {19.6172} \\
 ~ & PD-STD + IrisParseNet & {15.0566}  & {14.5395} & {14.3723}  & {14.2044}  & {15.3038} &  {14.9607}  \\
 ~ &PD-STD + DeepLabV3 & $\boldsymbol{14.3125}$  & $\boldsymbol{14.2262}$  & $\boldsymbol{13.9769}$  & $\boldsymbol{13.8321}$  & $\boldsymbol{14.1051}$ &  $\boldsymbol{14.4573}$ \\
 \hline
\end{tabular}
}
\caption{Results of the four networks of DeepLabV3, IrisParseNet and our proposed PD-STD-based networks, which are trained on images with Gaussian noise level $0.1$. 
Although the performance of DeepLabV3 and IrisParseNet improves a lot when training in the noisy image dataset,
our proposed PD-STD-based networks still perform better in all cases, especially in the compactness metric when segmenting images with Salt \& Pepper noises. }\label{table:trainOnnoisy}
\end{table}

% \begin{table}[]
% \centering
% \caption{Statistics of the PD-STD and DeepLabV3 algorithms on noisy Iris dataset}\label{table:Iris}
% \begin{tabular}{@{}cccc@{}}
% \toprule
%                 & \textbf{mIoU} &\textbf{dice} & \textbf{compactness}  \\ \midrule
% \textbf{PD-STD(depth = 100 )} & 0.6178    & 0.7014    & 13.3170                         \\
% \textbf{PD-STD(depth = 200)} & 0.5440    & 0.6567    & 12.9271                         \\
% \textbf{DeepLabV3}          & 0.4612    & 0.5337    & 21.3442                         \\ \bottomrule
% \end{tabular}
% \end{table}

% \begin{figure}[]
%     \centering
%     \includegraphics[width=10cm]{Figures/iris_compare2.png}
%     \caption{Results of different methods on noisy images}
%     \label{fig:iris}
% \end{figure}    

\subsubsection{Experiments on Fundus dataset}
% \paragraph{}\  
% In this section, we add $0.1, 0.07, 0.05, 0.01$ Gaussian noise and $0.001$ salt and pepper noise on Fundus testing dataset \cite{fundusDataset} on which the performance of PD-STD algorithm is evaluated.  There are $311$ training images and $174$ testing images in this dataset.  For fair comparison, we use the same parameters in the training process for both models.  The training images are pre-processed with crop size $ 300 \times 300$.  For this dataset, the batch size of the training is 20 and the training epoch is 200. 
More experiments on the image dataset of Fundus \cite{fundusDataset} are given in this section, 
which comprises $311$ training images and $174$ testing images.
Segmenting the optic disc region is essential for fundus image analysis. However, the segmentation result is often affected by existing blood vessels and a noisy imaging condition. 
In our experiments, the training images are cropped to the size of $300 \times 300$, 
and we set the training batch size to $20$, the training epoch to $200$, $\epsilon = 0.01$, and the iteration number to $50$. 
% In this section, we aim to evaluate the performance of the Fundus dataset \cite{fundusDataset}, which comprises $311$ training images and $174$ testing images. 
Both DeepLabV3 and the proposed PD-STD-DeepLabV3 are trained on the clean Fundus dataset, and tested on the images with different noise types and levels,
such as Gaussian noise levels of $0.01$, $0.05$, $0.07$, $0.1$, and Salt \& Pepper noise $0.01$.
The experiment results obtained by the baseline network DeepLabV3 and our proposed PD-STD+DeepLabV3 are presented in Tab. \ref{table:trainOnCleanfundus}.  
The proposed PD-STD+DeepLabV3 consistently outperforms DeepLabV3 in terms of IoU, Dice, and Compactness metrics across all noise levels. When tested on the images with a Gaussian noise level of $0.1$, the proposed PD-STD+DeepLabV3 significantly improves the accuracy of the segmentation results in terms of IoU and Dice. Fig. \ref{fig:fundus} visually exhibits the superior performance of our proposed PD-STD+DeepLabV3.
% demonstrates that the segmentation results obtained by our method exhibit superior performance.  

\begin{table}[t!]
\centering

\resizebox{\textwidth}{20mm}{
\begin{tabular}{@{}c|c|c|cccc|c@{}}
\hline
    & Method &Clean & \multicolumn{4}{c|}{Gaussian}  &  Salt \& Pepper \\ \hline
Noise level    &        &  0    &0.01 & 0.05 & 0.07 & 0.1 &     0.001   \\
                \hline
 \multirow{2}*{IoU}& DeepLabV3 & {0.9028} &  {0.9028} & {0.8648}  & {0.8254}  &{0.7681}  & {0.8220}    \\
 ~ &PD-STD + DeepLabV3 & $\boldsymbol{0.9061}$  & $\boldsymbol{0.9066}$  &  $\boldsymbol{0.8764}$  & $\boldsymbol{0.8540}$&$\boldsymbol{0.8264}$  & $\boldsymbol{0.8329}$    \\
\hline
  \multirow{2}*{Dice}& DeepLabV3 & {0.9482}  & {0.9483} & {0.9263} & {0.9028} & {0.8673} & {0.9008}   \\
 ~ &PD-STD + DeepLabV3 & $\boldsymbol{0.9500}$  & $\boldsymbol{0.9503}$ & $\boldsymbol{0.9331}$ & $\boldsymbol{0.9200}$ &  $\boldsymbol{0.9035}$  & $\boldsymbol{0.9076}$   \\
  \hline
  \multirow{2}*{Compactness}& DeepLabV3 & {14.3698}  & {14.3200}  & {14.7309} & {15.5674} &  {17.2642} & {15.6184}    \\
 ~ &PD-STD + DeepLabV3 & $\boldsymbol{14.0200}$  & $\boldsymbol{14.0228}$ & $\boldsymbol{14.1052}$ &  $\boldsymbol{14.2661}$ &  $\boldsymbol{14.4979}$ & $\boldsymbol{14.5199}$    \\
 \hline
\end{tabular}
}
\caption{DeepLabV3 and the proposed PD-STD+DeepLabV3 are trained on a clean fundus dataset. Clearly, the segmentation accuracy of DeepLabV3 quickly worsens as the noise level of test images gets bigger; in contrast, the proposed PD-STD+DeepLabV3 is not affected much by image noise levels.
}\label{table:trainOnCleanfundus}
\end{table}

% \begin{table}[]
% \centering
% \caption{Statistics of the PD-STD and DeepLabV3 algorithms on noisy Fundus dataset}\label{table:fundus}
% \begin{tabular}{@{}cccc@{}}
% \toprule
%                 & \textbf{mIoU} &\textbf{dice} & \textbf{compactness}  \\ \midrule
% \textbf{PD-STD($\epsilon = 0.01$ )} & 0.8147    & 0.8940    & 14.0875                         \\
% \textbf{PD-STD($\epsilon = 0.005$ )}& 0.7969    & 0.8822    & 14.0876                         \\
% \textbf{PD-STD($\epsilon = 0.001$ )}& 0.7848    & 0.8742    & 14.0222                         \\
% \textbf{DeepLabV3}          & 0.7071    & 0.8228    & 48.4696                         \\ \bottomrule
% \end{tabular}
% \end{table}

\begin{figure}[t!]
    \centering
    \includegraphics[width=0.8\textwidth]{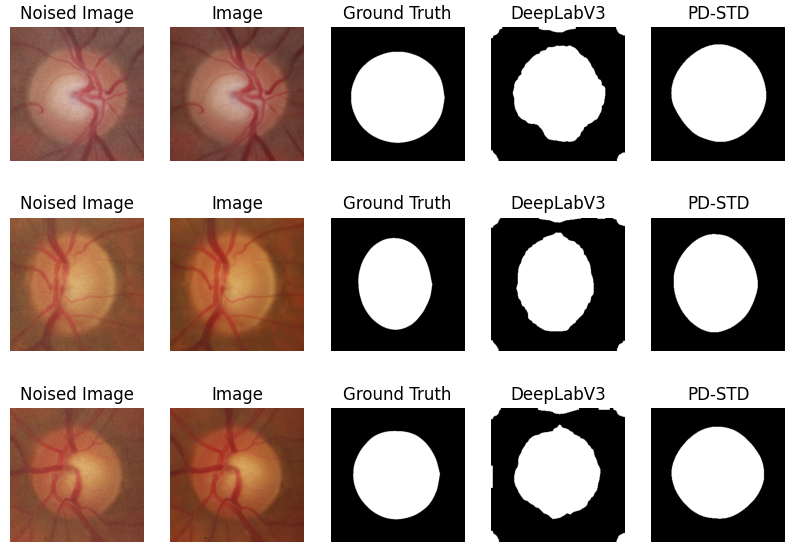}
    \caption{This pictures of this figure show some test results, obtained by the DeepLabV3 and PD-STD+DeepLabV3, on the images with Gaussian noise level $0.05$. The proposed PD-STD-DeepLabV3 obtains better segmentation results than DeepLabV3, with the correct shapes of optic disc which are nearly the same as the ground-truth results.}
    \label{fig:fundus}
\end{figure}

\begin{table}[t!]
\centering

% \begin{threeparttable}
\resizebox{\textwidth}{28mm}{
\begin{tabular}{@{}c|c|ccccccccc@{}}
\hline
    & {}  & \multicolumn{9}{c}{Noise level of test image dataset}   \\ \hline
    &    {Methods}  & {clean} &${0.1}$ & ${0.2}$ & ${0.3}$ & ${0.4}$ & ${0.5}$ & ${0.6}$ & ${0.7}$ & ${0.8}$   \\
                \hline
 \multirow{4}*{IoU}  & {DeepLabV3} & ${0.5389}$  & ${0.7936}$  & ${0.8545}$  & ${0.8639}$ & ${0.8689}$  & ${0.8666}$ &${0.8603}$ &${0.8561}$   &${0.8323}$  \\
 ~& {PD-STD+DeepLabV3}  & ${0.7159}$  & ${0.8133}$  & ${0.8493}$  & ${0.8656}$ & ${0.8702}$  & ${0.8793}$ &${0.8728}$ &${0.8633}$   &${0.8403}$ \\
 ~& {DeepLabV3*} & ${0.8706}$  & ${0.8744}$  & ${0.8772}$  & ${0.8837}$ & ${0.8867}$  & ${0.8876}$ &${0.8874}$ &${0.8835}$   &${0.8739}$  \\
 ~& {PD-STD+DeepLabV3*} & $\boldsymbol{0.8800}$  & $\boldsymbol{0.8836}$  & $\boldsymbol{0.8866}$  & $\boldsymbol{0.8903}$ & $\boldsymbol{0.8917}$  & $\boldsymbol{0.8894}$ &$\boldsymbol{0.8898}$ &$\boldsymbol{0.8868}$   &$\boldsymbol{0.8763}$  \\
 
 \hline
  \multirow{4}*{Dice}  & {DeepLabV3} & ${0.6671}$  & ${0.8660}$  & ${0.9081}$  & ${0.9141}$ & ${0.9179}$  & ${0.9153}$ &${0.9099}$ &${0.9098}$  &${0.8897}$ \\
 ~& {PD-STD+DeepLabV3} & ${0.8052}$  & ${0.8767}$  & ${0.9024}$  & ${0.9142}$ & ${0.9190}$  & ${0.9248}$ &${0.9212}$ &${0.9151}$  &${0.8970}$   \\
 ~& {DeepLabV3*} & ${0.9186}$  & ${0.9209}$  & ${0.9233}$  & ${0.9274}$ & ${0.9295}$  & ${0.9300}$ &${0.9304}$ &${0.9279}$  &${0.9215}$ \\
 ~& {PD-STD+DeepLabV3*} & $\boldsymbol{0.9241}$  & $\boldsymbol{0.9269}$  & $\boldsymbol{0.9289}$  & $\boldsymbol{0.9312}$ & $\boldsymbol{0.9322}$  & $\boldsymbol{0.9309}$ &$\boldsymbol{0.9314}$ &$\boldsymbol{0.9299}$  &$\boldsymbol{0.9231}$ \\
  \hline
  \multirow{4}*{Comp.}  & {DeepLabV3}  & ${50.8273}$  & ${24.4920}$  & ${17.9213}$  & ${16.7283}$ & ${15.6044}$  & ${15.3893}$ &${15.4148}$ &${15.7572}$ &${17.7720}$  \\
 ~& {PD-STD+DeepLabV3} & ${16.4577}$  & ${14.6549}$  & $\boldsymbol{14.2005}$  & ${14.1541}$ & ${14.1028}$  & $\boldsymbol{13.9716}$ &${14.0589}$ &${14.0268}$ &$\boldsymbol{13.9119}$  \\
 ~& {DeepLabV3*} & ${16.2839}$  & ${16.7775}$  & ${16.4507}$  & ${15.6798}$ & ${15.2507}$  & ${14.8319}$ &${14.8209}$ &${14.9177}$ &${15.0247}$  \\
 ~& {PD-STD+DeepLabV3*} & $\boldsymbol{14.4574}$  & $\boldsymbol{14.3944}$  & ${14.2638}$  & $\boldsymbol{14.0775}$ & $\boldsymbol{14.0686}$  & ${13.9879}$ &$\boldsymbol{13.9769}$ &$\boldsymbol{13.9992}$ &${13.9606}$  \\
 \hline
\end{tabular}}
\caption{Comparison between the direct training strategy and the progressive training strategy:
% on a dataset with a Gaussian noise level of $0.8$ and progressive training with the highest Gaussian noise level of $0.8$.  Methods employing progressive training at the highest Gaussian noise level of $0.8$ are marked with '*'.
the networks of DeepLabV3 and our proposed PD-STD+DeepLabV3 are trained directly by the image dataset of a high Gaussian noise level of $0.8$; and the trained networks of DeepLabV3 and PD-STD+DeepLabV3 by the progressive training strategy are denoted by DeepLabV3* and PD-STD+DeepLabV3* correspondingly.
Clearly, the progressive training strategy does improve the performance of the trained networks significantly in terms of of IoU, Dice and Compactness, and the networks of DeepLabV3* and PD-STD+DeepLabV3*, trained in the progressive way, perform much more reliably on test images across different noise levels.
}\label{table:proVSnopro}
% \begin{tablenotes}
%     \small
%     \item DeepLabV3*: We use the progressive training method with the highest Gaussian noise level of $0.8$ to train the DeepLabV3 network
%     \item PD-STD*: We use the progressive training method with the highest Gaussian noise level of $0.8$ to train our PD-STD network.
% \end{tablenotes}
% \end{threeparttable}
\end{table}

\subsection{Progressive Training Strategy}
%\paragraph{}\ 
In this work, a progressive training strategy is also used to train the neural networks and update their parameters progressively from the image dataset of low noise level to the one with high noise level. 
As shown in \cite{tai2023pottsmgnet}, such progressive training strategy improves the robustness and reliability of trained neural networks, compared to the direct training strategy which trains neural networks directly on the image dataset with a high level of noise.
% The progressive training strategy used in \cite{tai2023pottsmgnet} involves training a neural network on a dataset with a specific noise level and then using the trained parameters to initialize the network for training on the next dataset with a higher noise level. 
% This process is repeated iteratively, with each subsequent dataset having a progressively higher noise level than the previous one.  
A comprehensive comparison of the direct training strategy and the introduced progressive training strategy is shown in Tab. \ref{table:proVSnopro}:
the networks of DeepLabV3 and our proposed PD-STD+DeepLabV3 are trained by the two training strategies at a high Gaussian noise level of $0.8$, respectively; the trained networks of DeepLabV3 and PD-STD+DeepLabV3 by the progressive training strategy are denoted by DeepLabV3* and PD-STD+DeepLabV3* correspondingly.

\begin{figure*}[t!]
  \centering 
  \begin{minipage}[b]{0.9\textwidth} 
  {
    \begin{minipage}[b]{0.20\textwidth}
      Noisy Image \   \  \vspace{60pt} \\
      Ground Truth\  \vspace{65pt} \\
      DeepLabV3\ \   \vspace{50pt} \\
      PD-STD\\+DeepLabV3\  \vspace{60pt} \\
      DeepLabV3*  \  \vspace{60pt} \\
      PD-STD\\+DeepLabV3* \   \\
    \end{minipage}
  }
  \hfill
   \subfloat[]{
    \begin{minipage}[b]{0.22\textwidth}
      \centering
      \includegraphics[width=\linewidth]{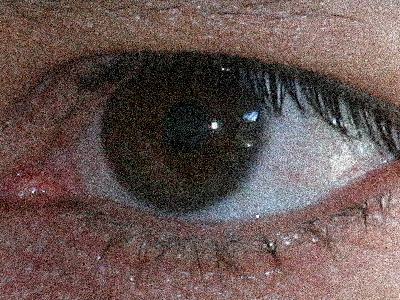}\vspace{5pt}
      \includegraphics[width=\linewidth]{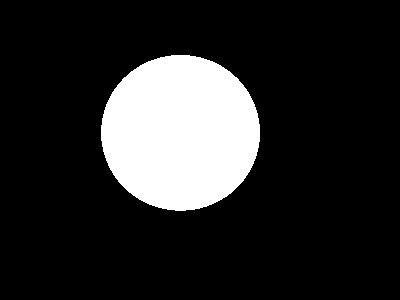}\vspace{5pt}
      \includegraphics[width=\linewidth]{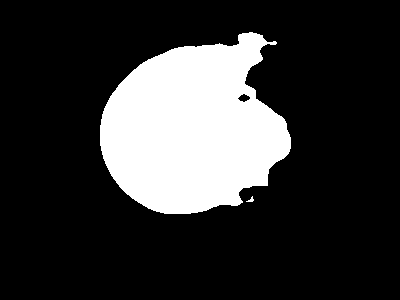}\vspace{5pt}
      \includegraphics[width=\linewidth]{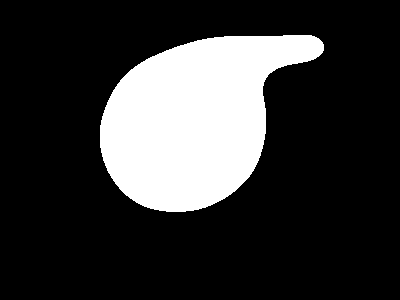}\vspace{5pt}
      \includegraphics[width=\linewidth]{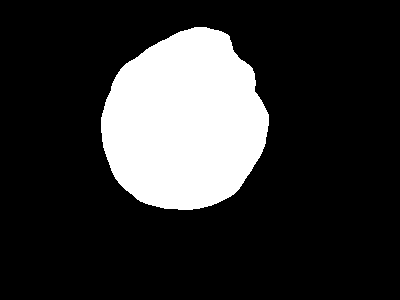}\vspace{5pt}
      \includegraphics[width=\linewidth]{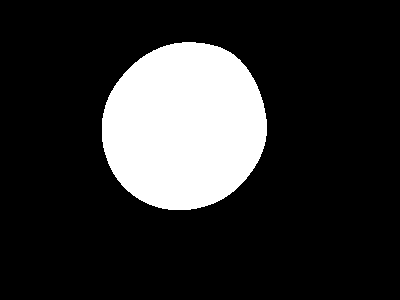}
    \end{minipage}
  }
  \hfill
   \subfloat[]{
    \begin{minipage}[b]{0.22\textwidth}
      \centering
      \includegraphics[width=\linewidth]{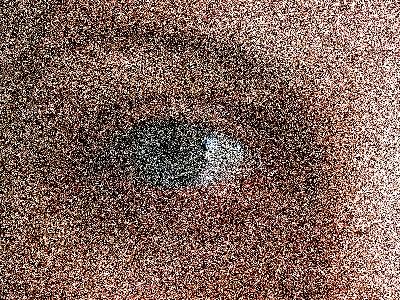}\vspace{5pt}
      \includegraphics[width=\linewidth]{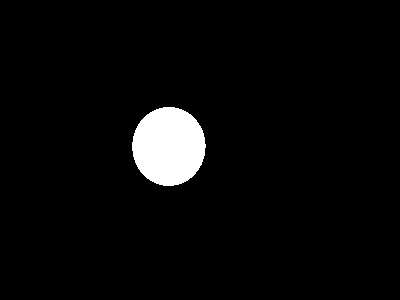}\vspace{5pt}
      \includegraphics[width=\linewidth]{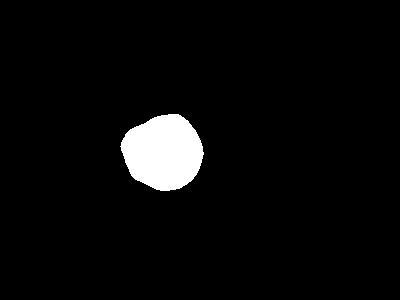}\vspace{5pt}
      \includegraphics[width=\linewidth]{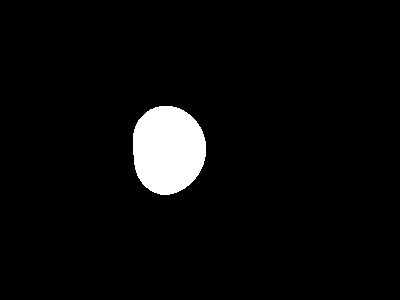}\vspace{5pt}
      \includegraphics[width=\linewidth]{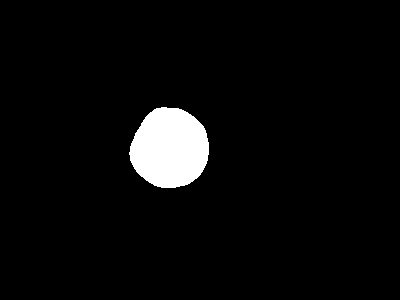}\vspace{5pt}
      \includegraphics[width=\linewidth]{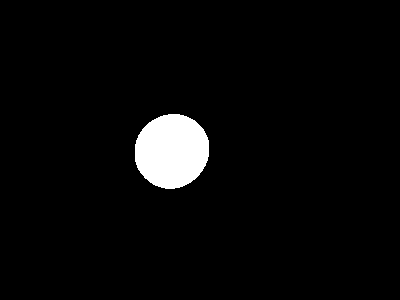}
    \end{minipage}
  }
  \hfill
    \subfloat[]{
    \begin{minipage}[b]{0.22\textwidth}
      \centering
      \includegraphics[width=\linewidth]{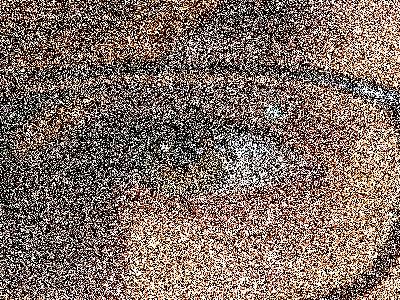}\vspace{5pt}
      \includegraphics[width=\linewidth]{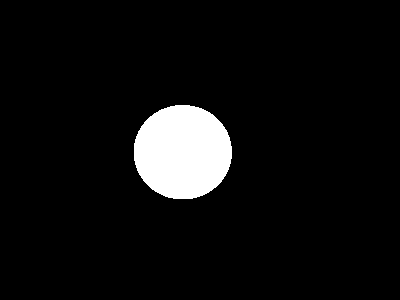}\vspace{5pt}
      \includegraphics[width=\linewidth]{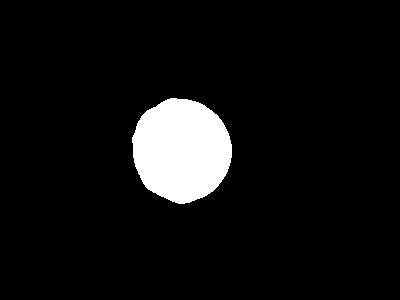}\vspace{5pt}
      \includegraphics[width=\linewidth]{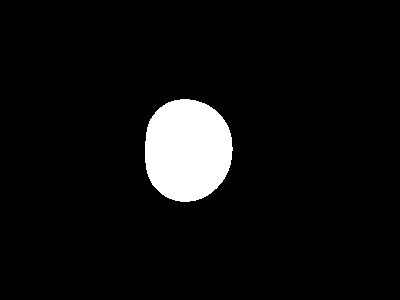}\vspace{5pt}
      \includegraphics[width=\linewidth]{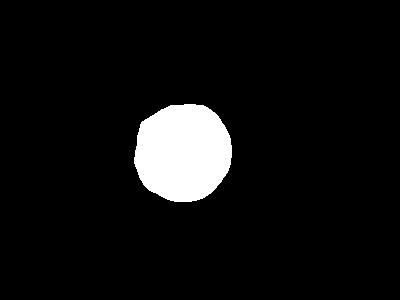}\vspace{5pt}
      \includegraphics[width=\linewidth]{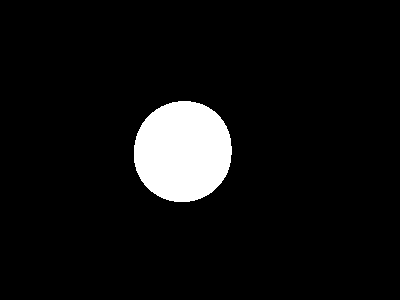}
    \end{minipage}
  }
  \end{minipage}
  \vfill
  \caption{
  Three segmentation expamples demonstrated in this figure show that the progressive training strategy largely improves the performance of the trained networks comparing to the direct training strategy.
  % Segmentation results predicted using direct training on a dataset with a Gaussian noise level of $0.8$ and progressive training with the highest Gaussian noise level of $0.8$. 
  Gaussian noise levels of the test images in (a), (b), and (c) are $0.1$, $0.5$, and $0.8$, respectively. Obviously, the networks trained by the direct training strategy at a high noise level of $0.8$ perform much worse on the test image with a low noise level.
  }\label{fig:proVSnoprofigures}
\end{figure*}

It is easy to see that the progressive training strategy does improve the performance of trained networks significantly in terms of of IoU, Dice and Compactness; and the networks of DeepLabV3* and PD-STD+DeepLabV3*, which are trained in the progressive way, perform much more robustly and reliably on test images of different noise levels.
Fig. \ref{fig:proVSnoprofigures} clearly illustrates some examples: the networks trained by the direct training strategy at a high noise level of $0.8$ perform much worse on the test image with a low noise level.

\begin{table}[t!]
\centering
\resizebox{\textwidth}{43mm}{
\begin{tabular}{@{}cccccccccc@{}}
\hline 
    {\diagbox[width=2.5cm]{test\\ noise level}{training noise\\ level}}  &{} &${0.1}$ & ${0.2}$ & ${0.3}$ & ${0.4}$ & ${0.5}$ & ${0.6}$ & ${0.7}$ & ${0.8}$   \\
  \hline
  \multirow{2}*{clean} & {DeepLabV3} & ${14.57}$    & ${15.26}$    & ${15.96}$  & ${17.30}$  & ${17.67}$   & ${16.52}$  & ${16.99}$  & ${16.29}$  \\
  ~& {PD-STD+DeepLabV3} & $\boldsymbol{13.89}$  & $\boldsymbol{14.10}$    & $\boldsymbol{14.06}$  &$ \boldsymbol{13.96}$  &$ \boldsymbol{14.13}$   &$ \boldsymbol{14.23}$  &$ \boldsymbol{14.56}$  &$ \boldsymbol{14.47}$  \\
  \cline{1-10}
 ~ \multirow{2}*{0.1} & {DeepLabV3} & ${14.38}$  & ${15.00}$  & ${15.79}$  & ${16.95}$  & ${17.65}$  & ${16.85}$ & ${18.44}$  & ${16.79}$   \\
 ~& {PD-STD+DeepLabV3} & $\boldsymbol{13.87}$  & $\boldsymbol{14.05}$    & $\boldsymbol{14.16}$  &$\boldsymbol {14.03}$  &$\boldsymbol {14.16}$  &$\boldsymbol {14.30}$ &$ \boldsymbol{14.32}$  &$ \boldsymbol{14.40}$   \\
 \cline{1-10}
 ~ \multirow{2}*{0.2} & {DeepLabV3} & ${23.11}$  & ${14.64}$ & ${14.67}$  & ${15.74}$  & ${16.55}$  & ${16.69}$ & ${17.17}$  & ${16.46}$   \\
 ~& {PD-STD+DeepLabV3} & $\boldsymbol{15.71}$  & $\boldsymbol{13.81}$    & $\boldsymbol{14.01}$  &$\boldsymbol {13.97}$  &$\boldsymbol {14.05}$  & $\boldsymbol{14.09}$ &$\boldsymbol {14.09}$  &$\boldsymbol {14.27}$   \\
 \cline{1-10}
 ~ \multirow{2}*{0.3} & {DeepLabV3} & ${-}$  & ${22.34}$  & ${14.55}$  & ${14.88}$  & ${15.44}$  & ${15.68}$ & ${15.72}$  & ${15.69}$   \\
 ~& {PD-STD+DeepLabV3}  & ${-}$   & $\boldsymbol{14.29}$    & $\boldsymbol{13.93}$  & $\boldsymbol{13.89}$  & $\boldsymbol{13.96}$  & $\boldsymbol{14.02}$ &$ \boldsymbol{14.10}$  &$ \boldsymbol{14.09}$   \\
 \cline{1-10}
 ~ \multirow{2}*{0.4} & {DeepLabV3} & ${-}$  & ${43.80}$ & ${16.10}$  & ${14.51}$  & ${14.84}$  & ${14.87}$ & ${15.07}$  & ${15.26}$  \\
 ~& {PD-STD+DeepLabV3}  & ${-}$        & $\boldsymbol{13.64}$    & $\boldsymbol{14.01}$  & $\boldsymbol{13.83}$  & $\boldsymbol{13.95}$  & $\boldsymbol{13.95}$ & $\boldsymbol{13.92}$  &$ \boldsymbol{14.14}$  \\
 \cline{1-10}
 ~ \multirow{2}*{0.5} & {DeepLabV3} & ${-}$  & ${-}$ & ${25.04}$  & ${15.92}$  & ${14.57}$  & ${14.99}$ & ${14.88}$  & ${14.84}$   \\
 ~& {PD-STD+DeepLabV3}  & ${-}$        & ${-}$     & $\boldsymbol{14.35}$  & $\boldsymbol{13.81}$  & $\boldsymbol{13.92}$  & $\boldsymbol{13.86}$ & $\boldsymbol{13.92}$  & $\boldsymbol{14.00}$   \\
 \cline{1-10}
 ~ \multirow{2}*{0.6} & {DeepLabV3} & ${-}$  & ${-}$ & ${-}$  & ${23.84}$  & ${16.50}$  & ${15.27}$ & ${14.70}$  & ${14.83}$   \\
 ~& {PD-STD+DeepLabV3}  & ${-}$        & ${-}$     & ${-}$  & $\boldsymbol{13.37}$  & $\boldsymbol{14.05}$  & $\boldsymbol{13.99}$ & $\boldsymbol{14.01}$  & $\boldsymbol{13.99}$   \\
 \cline{1-10}
 ~ \multirow{2}*{0.7} & {DeepLabV3} & ${-}$  & ${-}$ & ${-}$  & ${-}$  & ${22.38}$  & ${15.32}$ & ${14.81}$  & ${14.93}$    \\
 ~& {PD-STD+DeepLabV3}  & ${-}$        & ${-}$     & ${-}$  & ${-}$  & $\boldsymbol{14.02}$  & $\boldsymbol{13.84}$ & $\boldsymbol{13.91}$  & $\boldsymbol{14.01}$    \\
 \cline{1-10}
 ~ \multirow{2}*{0.8} & {DeepLabV3} & ${-}$  & ${-}$ & ${-}$  & ${-}$  & ${32.61}$  & ${20.58}$ & ${15.91}$  & ${15.03}$   \\
 ~& {PD-STD+DeepLabV3}  & ${-}$        & ${-}$     & ${-}$  & ${-}$  & $\boldsymbol{13.77}$  & $\boldsymbol{13.88}$ & $\boldsymbol{13.91}$  & $\boldsymbol{13.97}$   \\
 \hline
\end{tabular}}
\caption{
The progressive training strategy is applied for the Iris dataset from the image noise level of $0.1$ to $0.8$ gradually at a step-size $0.1$. 
Given the circular shape of Iris, the compactness of the optimal image segmentation result tends to be $4\pi \approx 12.56$; hence, the compactness value closer to $12.56$ means a more circular segmentation region is obtained, i.e. better. Clearly, our proposed PD-STD+DeepLabV3 performs more accuractely and robustly than DeepLabV3 when trained on a fixed image noise level but tested on different image noise levels; also, the same when trained on different image noise levels but tested on a fixed image noise level. In all cases, our proposed PD-STD+DeepLabV3 achieves better results than DeepLabV3.
}\label{table:progress no STD}
\end{table}

The progressive training strategy is employed for the Iris dataset from the image noise level of $0.1$ to $0.8$ gradually at a step-size $0.1$ and its extensive results are shown in Tab. \ref{table:progress no STD}. 
Given the circular shape of Iris, the compactness of the optimal image segmentation result tends to be $4\pi \approx 12.56$; so the compactness value closer to $12.56$ means a more circular segmentation region is obtained, which means better. In view of this, our proposed PD-STD+DeepLabV3 performs better than DeepLabV3 in both accuracy and robustness when trained at a fixed image noise level but tested at different image noise levels; also, the same when trained at different image noise levels but tested at a fixed image noise level. In all cases, our proposed PD-STD+DeepLabV3 achieves better results than DeepLabV3.
Fig. \ref{fig:miouprogressive} (a) and (b) demonstrate the performance of two examples of such experiments, in terms of IoU, when trained at a fixed image noise level but tested on different image noise levels: with the help of the introduced PD-STD block, i.e. incorporating a compact shape prior information, our proposed PD-STD+DeepLabV3 can obtain more accurate results than DeepLabV3 when trained at a fixed image noise level of $0.5$ (a) and $0.7$ (b) but tested at different image noise levels; moreover,  
as (a) shown, when both networks are trained at the image noise level of $0.5$, DeepLabV3 achieves the result close to our proposed PD-STD+DeepLabV3 at test image noise level of $0.5$, 
but it performs much worse than PD-STD+DeepLabV3 at a high test image noise level of $0.8$; the proposed PD-STD+DeepLabV3 exhibits better robustness than DeepLabV3 in both (a) and (b), which enables PD-STD-based networks to properly reduce the influence of real-world high noise level and data variability.  
Fig. \ref{fig:trainOnProgressive} provides three illustrative examples which show that the proposed PD-STD network block can effectively incorporate proper shape information, thus ensuring more reasonable image segmentation results when noise and data variability exist.  

As the above experiments show, by training the networks on the image datasets with higher noise levels and progressively increasing the noise level during training, the networks can be gradually adapted to different noisy images and memorize image views of different noise levels, which hence enhances the networks' ability to handle image noise and improves their performance on datasets with various noise levels. 
The progressive training strategy is therefore essential to improve the trained networks' performance in accuracy and robustness across different image noise levels.
\begin{figure*}[t!]
  \centering 
  \begin{minipage}[b]{1.0\linewidth} 
  {
    \begin{minipage}[b]{0.5\linewidth}
      \centering
      % \subcaptionbox*{(a)\label{subfig:fig_a}}{
      % \includegraphics[width=\linewidth]{Figures/progressive/noisy05_font15.png}\vspace{5pt}
      % }
      \subcaptionbox*{(a)\label{subfig:fig_a}}{
      \includegraphics[width=\linewidth]{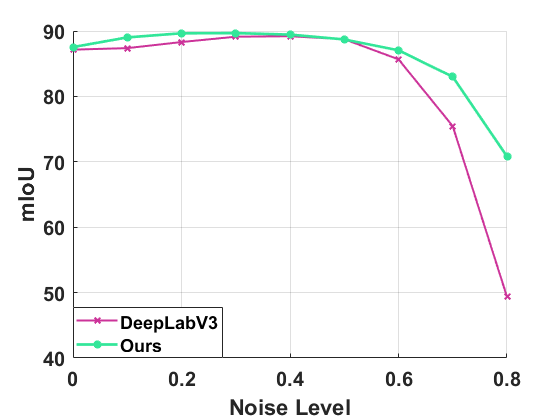}\vspace{5pt}
      }
    \end{minipage}
  }
  \hfill
   {
    \begin{minipage}[b]{0.5\linewidth}
      \centering
      % \subcaptionbox*{(b)\label{subfig:fig_b}}{
      % \includegraphics[width=\linewidth]{Figures/progressive/noisy07_font15.png}\vspace{5pt}
      % }
      \subcaptionbox*{(b)\label{subfig:fig_b}}{
      \includegraphics[width=\linewidth]{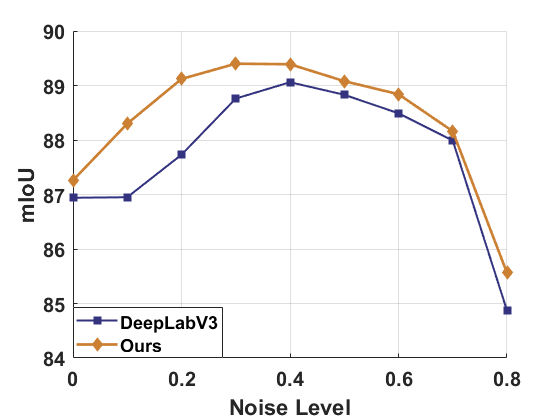}\vspace{5pt}
      }
    \end{minipage}
  }
  \end{minipage}
  \vfill
  \caption{The two graphs (a) and (b) show the performance of DeepLabV3 and our proposed PD-STD+DeepLabV3 on the test images with different noise levels, when the networks are trained progressively at the noise level of $0.5$ and $0.7$ respectively. Our proposed PD-STD+DeepLabV3 outperforms DeepLabV3 in all cases in terms of IoU, and demonstrates more robustly when there is a big difference in noise level between the training image data and the test image data.
  % IoU performance achieved through progressive training with the highest noise level $0.8$ on the Iris Dataset.  (a) and (b) correspond to models trained on a dataset with the highest Gaussian noise levels of $0.5$ and $0.7$, respectively. 
  }\label{fig:miouprogressive}
\end{figure*}

\begin{figure*}[t!]
  \centering 
  \begin{minipage}[b]{0.85\linewidth} 
  \subfloat[Noisy Image]{
    \begin{minipage}[b]{0.22\linewidth}
      \centering
      \includegraphics[width=\linewidth]
      {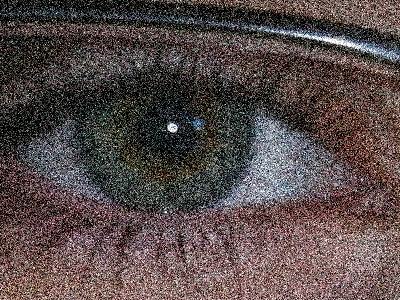}\vspace{5pt}
      \includegraphics[width=\linewidth]{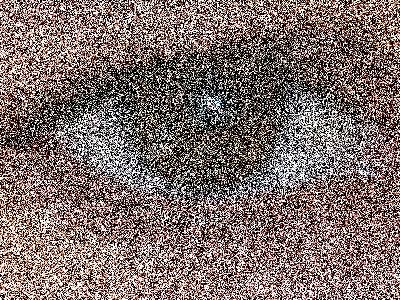}\vspace{5pt}
      \includegraphics[width=\linewidth]
      {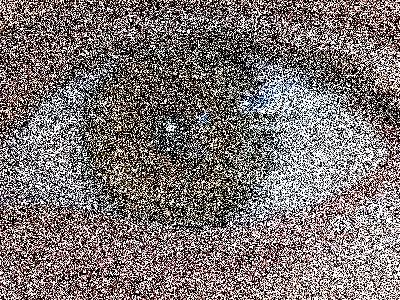}\vspace{5pt}
    \end{minipage}
  }
  \hfill
   \subfloat[Ground Truth]{
    \begin{minipage}[b]{0.22\linewidth}
      \centering
      \includegraphics[width=\linewidth]
      {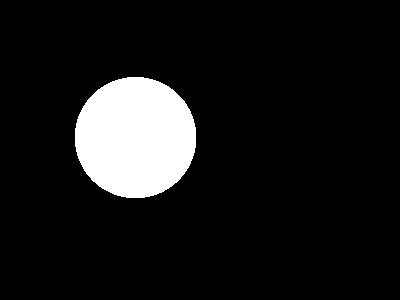}\vspace{5pt}
      \includegraphics[width=\linewidth]{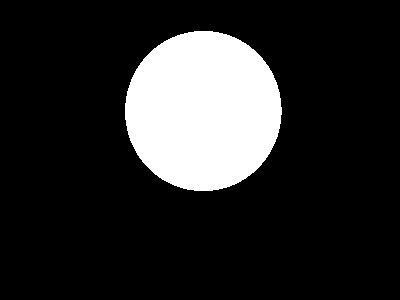}\vspace{5pt}
      \includegraphics[width=\linewidth]
      {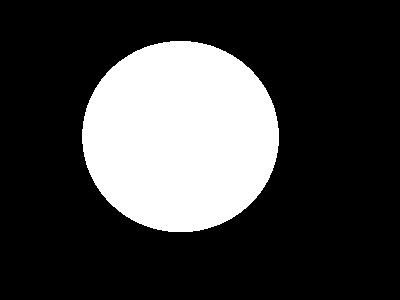}\vspace{5pt}
    \end{minipage}
  }
  \hfill
    \subfloat[DeepLabV3]{
    \begin{minipage}[b]{0.22\linewidth}
      \centering
      \includegraphics[width=\linewidth]{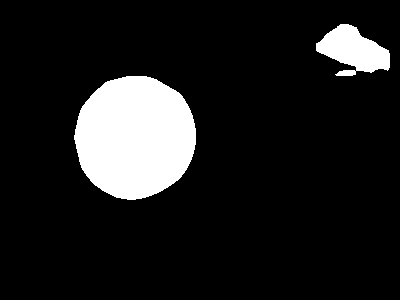}\vspace{5pt}
      \includegraphics[width=\linewidth]{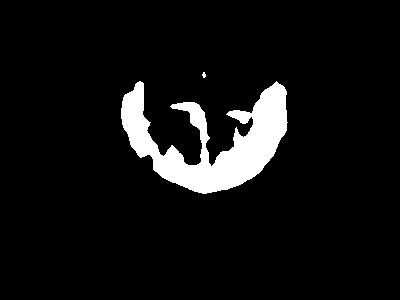}\vspace{5pt}
      \includegraphics[width=\linewidth]
      {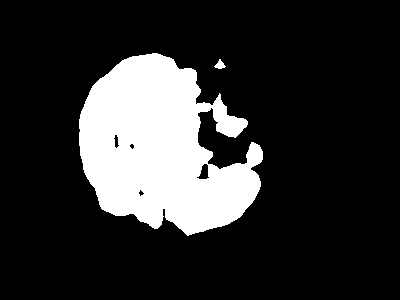}\vspace{5pt}
    \end{minipage}
  }
  \hfill
    \subfloat[Ours]{
    \begin{minipage}[b]{0.22\linewidth}
      \centering
      \includegraphics[width=\linewidth]{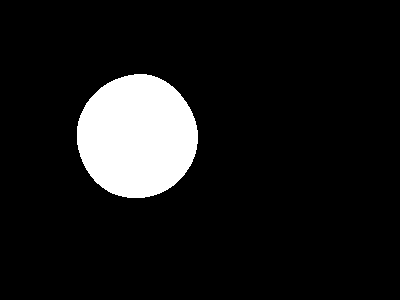}\vspace{5pt}
      \includegraphics[width=\linewidth]{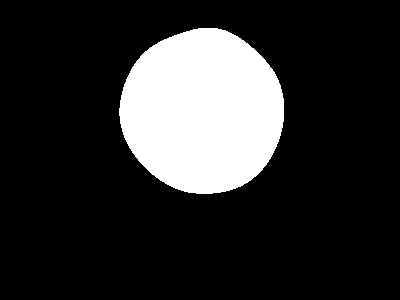}\vspace{5pt}
      \includegraphics[width=\linewidth]
      {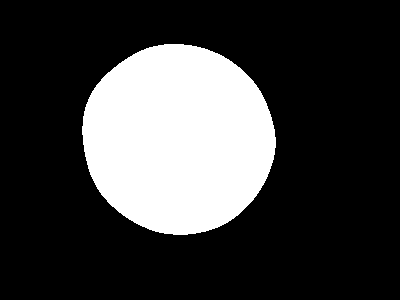}\vspace{5pt}
    \end{minipage}
  }
  \end{minipage}
  \vfill
  \caption{Segmentation results of progressive training with noise level varying from $0.1$ to $0.8$. Models were trained and tested at different noise levels, with the noisy training images having noise levels of $0.5$, $0.5$, and $0.7$ from top to bottom. The segmentation results for DeepLabV3 and our method correspond to models tested on noisy testing images with noise levels of $0.2$, $0.7$, and $0.8$.} \label{fig:trainOnProgressive}
\end{figure*}

\section{Conclusion} \label{sec: Conclusion}
In this paper, we proposed two novel algorithms PD-TD and PD-STD to solve the challenging image segmentation problem with a high-order shape-compactness prior, which essentially evaluates the ratio of squared perimeter to area.  
The new algorithms are based on the new  
primal-dual model, which is equivalent to the studied optimization problem, outperform existing methods in numerical simplicity and effectiveness. 
% Unlike existing methods, our method introduced the new equivalent primal-dual model to the studied optimization problem to avoid heavy iterations and reach the local minimum easier and faster.  
Meanwhile, a new PD-STD block is introduced to replace the often-used sigmoid layer of the backbone DNNs, which properly integrates the shape-compactness information into the neural network and enforces compact regions in segmentation results.
% Meanwhile, based on the variational explanation of the sigmoid layer in 
% DNNs, we have taken efforts to integrate the compactness term into the data-driven DNN models and proposed a PD-STD  block to replace the sigmoid layer.  
% This block can help DNNs generate more compact segmentation results with smoother boundaries.  
% Moreover, our method is more robust to noise.  We added Gaussian noise and salt-and-pepper noise to the Iris and Fundus datasets and applied our PD-STD block on the backbone network DeepLabV3 on these datasets.  
Extensive experiments, especially on highly noisy image datasets, show that the proposed PD-STD-based neural networks significantly outperform the state-of-the-art DNNs in both robustness and accuracy. 
Such PD-STD block can also be applied to many other DNN models, besides DeepLabV3 and IrisParseNet used in this work.

\section*{Acknowledgements}
We acknowledge the following funding sources that supported this work. Dr. Hao Liu was partially supported by the Natural Science Foundation of China (Grant No. 12201530) and the HKRGC ECS (Grant No. 22302123). Professor Jing Yuan acknowledges the support from the National Natural Science Foundation of China (NSFC) under Grant No. 61877047, as well as the Distinguished Professorship Start-up Funding of Zhejiang Normal University (No. YS304022908).  Professor Xue-cheng Tai was partially supproted by the NORCE Kompetanseoppbygging program.

\bibliographystyle{elsarticle-num} 
\bibliography{bibfile}

% \bibliography{bibfile} 
% \bibliographystyle{abbrv}
\end{document}